\documentclass[runningheads]{llncs}
\usepackage{graphicx}
\usepackage{subfigure}
\usepackage{amsmath,amssymb} 
\usepackage{tabularx}
\usepackage{multirow}
\usepackage{color}
\usepackage{colortbl}
\usepackage{makecell}

\newcommand{\figref}[1]{Fig. \ref{#1}}
\newcommand{\tabref}[1]{Table \ref{#1}}

\newcommand{\secref}[1]{Sec. \ref{#1}}

\definecolor{srcolor}{rgb}{1,0,0}

\definecolor{dbcolor}{rgb}{0,0,1}

\makeatletter
\def\hlinewd#1{
	\noalign{\ifnum0=`}\fi\hrule \@height #1 \futurelet
	\reserved@a\@xhline}

\begin{document}
\pagestyle{headings}
\mainmatter

\title{PARN: Pyramidal Affine Regression Networks for Dense Semantic Correspondence} 

\titlerunning{PARN: Pyramidal Affine Regression Networks}

\authorrunning{S.Jeon, S.Kim, D.Min, K.Sohn}

\author{Sangryul Jeon\textsuperscript{1}, Seungryong Kim\textsuperscript{1}, Dongbo Min\textsuperscript{2}, Kwanghoon Sohn\textsuperscript{1,}\thanks{Corresponding Author}}


\institute{\textsuperscript{1}Yonsei University\\
	\email{ \{cheonjsr,srkim89,khsohn\}@yonsei.ac.kr}\\
	\textsuperscript{2}Ewha Womans University\\
	\email{dbmin@ewha.ac.kr}
}

\maketitle

\begin{abstract}
This paper presents a deep architecture for dense semantic
correspondence, called pyramidal affine regression networks (PARN),
that estimates locally-varying affine transformation fields across images.
To deal with intra-class appearance and shape variations that commonly exist among different instances within the same object category,
we leverage a pyramidal model where affine transformation fields are progressively
estimated in a coarse-to-fine manner so that the smoothness
constraint is naturally imposed within deep networks. PARN estimates residual affine transformations at each level and composes them
to estimate final affine transformations. Furthermore, to
overcome the limitations of insufficient training data for semantic
correspondence, we propose a novel weakly-supervised training scheme that generates progressive supervisions by leveraging a correspondence consistency across image pairs. Our method is fully learnable in an
end-to-end manner and does not require quantizing infinite continuous affine transformation fields. To the best of our knowledge, it is the
first work that attempts to estimate dense affine
transformation fields in a coarse-to-fine manner within deep
networks. Experimental results demonstrate that PARN outperforms the
state-of-the-art methods for dense semantic correspondence on various
benchmarks.
\keywords{dense semantic correspondence, hierarchical graph model}
\end{abstract}

\section{Introduction}
Establishing dense correspondences across semantically similar images is essential for numerous computer vision and computational photography applications, such as
scene parsing, semantic
segmentation, and image editing \cite{HaCohen2011,Liu11,Kim13,Yang14,Zhou15}.

Unlike classical dense correspondence tasks such as stereo matching
\cite{taxonomy} or optical flow estimation \cite{butler} that have been
dramatically advanced, semantic correspondence estimation still
remains unsolved due to severe intra-class appearance and shape
variations across images. Several recent approaches \cite{fcss,ucn} have been
proposed by leveraging deep convolutional neural networks (CNNs),
providing satisfactory performances in capturing reliable matching
evidences under intra-class appearance variations. However,
they still consider geometric variations in just a limited manner such as those used for
stereo matching or optical flow estimation \cite{taxonomy,butler}. In some
approaches \cite{ucn,dctm}, more complex geometric variations such as scale
or rotation were addressed, but they seek the labeling solution from only a set of scales and/or rotations quantized within pre-defined
ranges. Recently, the discrete-continuous transformation matching
(DCTM) framework \cite{dctm} combined with the fully convolutional
self-similarity (FCSS) \cite{fcss} descriptor exhibits much improved
performance by estimating locally-varying affine transformation
fields on continuous and discrete domains in an alternative manner.
Although DCTM has shown the state-of-the-art performance in
dealing with non-rigid shape deformations, it is formulated with
handcrafted smoothness constraint model and optimization technique, and thus it cannot guarantee optimal results when the geometric variation is relatively large.
\begin{figure}[t!]
	\centering
	\renewcommand{\thesubfigure}{}
	\subfigure[]
	{\includegraphics[width=0.164\linewidth]{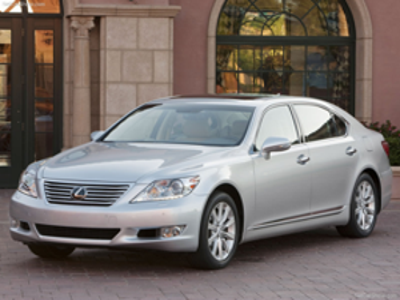}}\hfill
	\subfigure[]
	{\includegraphics[width=0.164\linewidth]{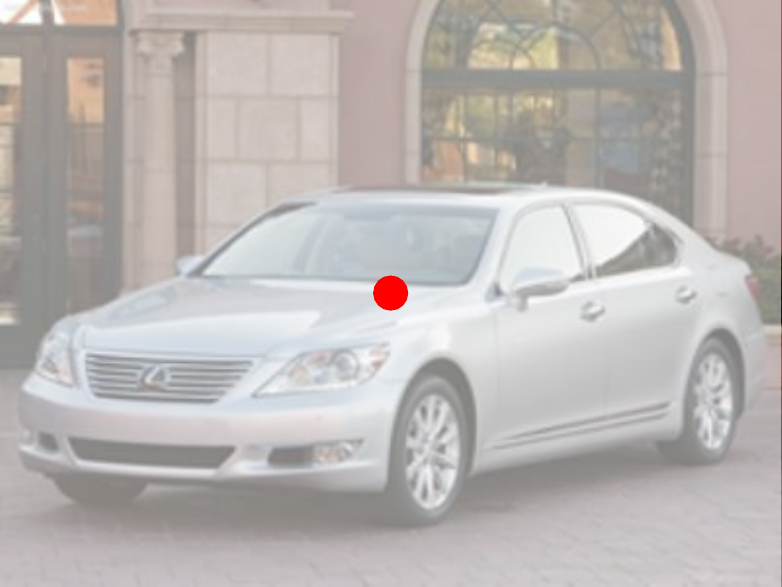}}\hfill
	\subfigure[]
	{\includegraphics[width=0.164\linewidth]{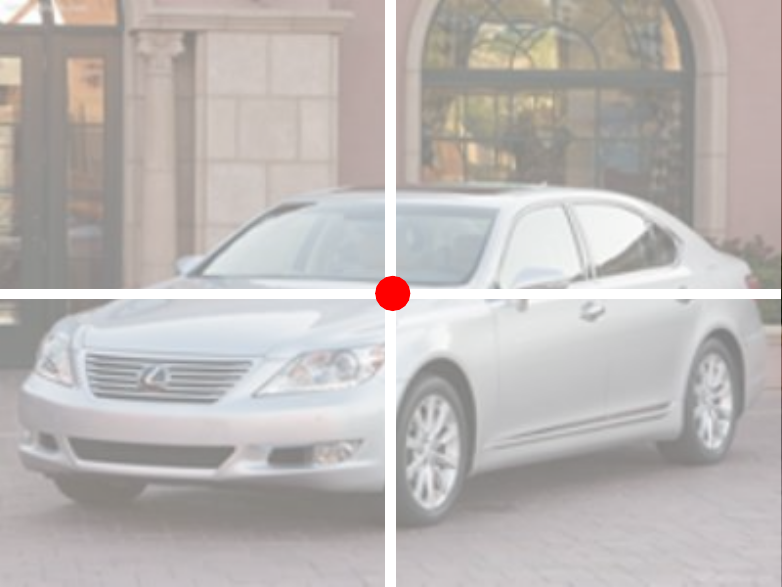}}\hfill
	\subfigure[]
	{\includegraphics[width=0.164\linewidth]{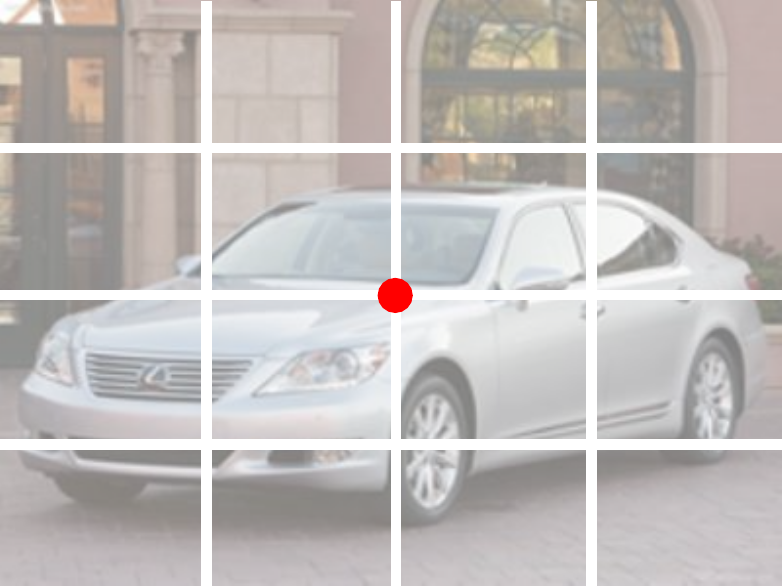}}\hfill
	\subfigure[]
	{\includegraphics[width=0.164\linewidth]{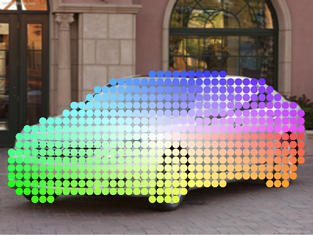}}\hfill
	\subfigure[]
	{\includegraphics[width=0.164\linewidth]{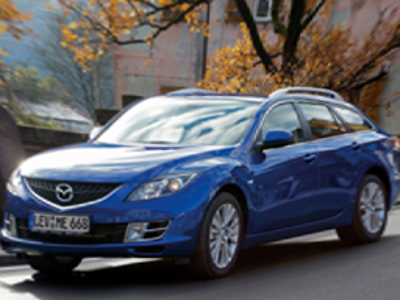}}\hfill
	\vspace{-21pt}
	\subfigure[(a)]
	{\includegraphics[width=0.164\linewidth]{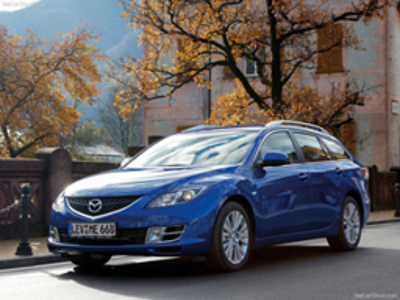}}\hfill
	\subfigure[(b)]
	{\includegraphics[width=0.164\linewidth]{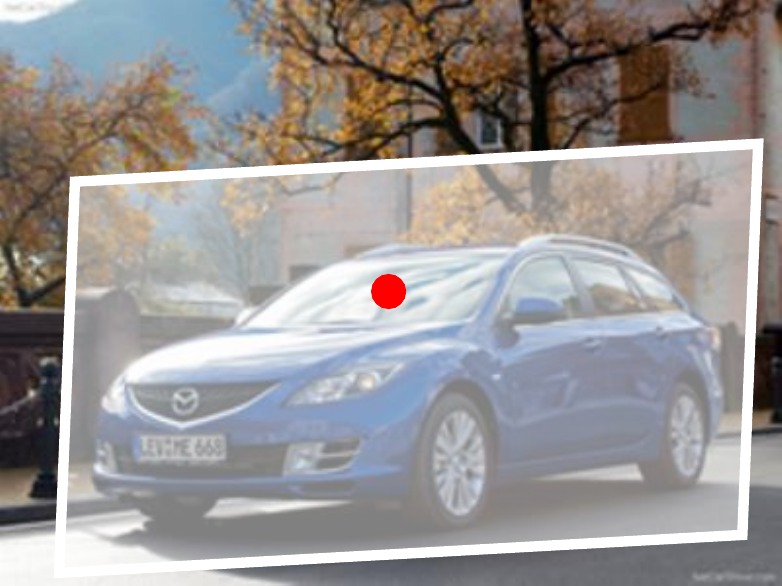}}\hfill
	\subfigure[(c)]
	{\includegraphics[width=0.164\linewidth]{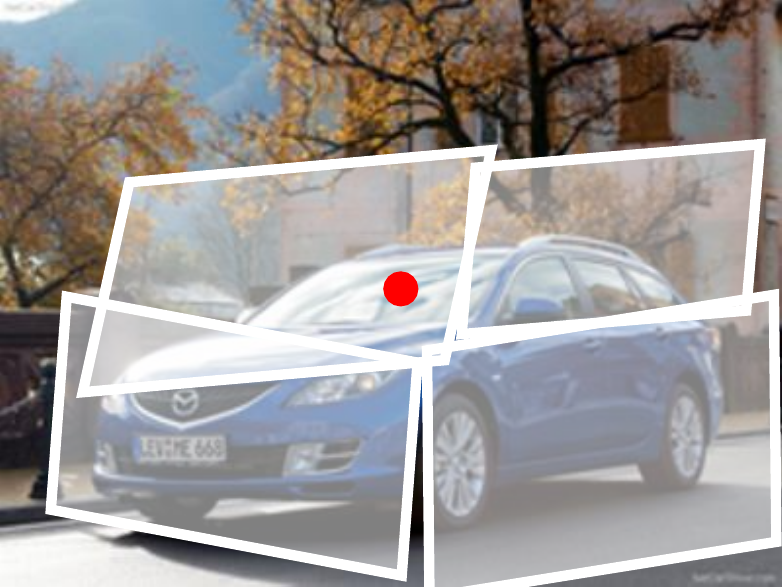}}\hfill
	\subfigure[(d)]
	{\includegraphics[width=0.164\linewidth]{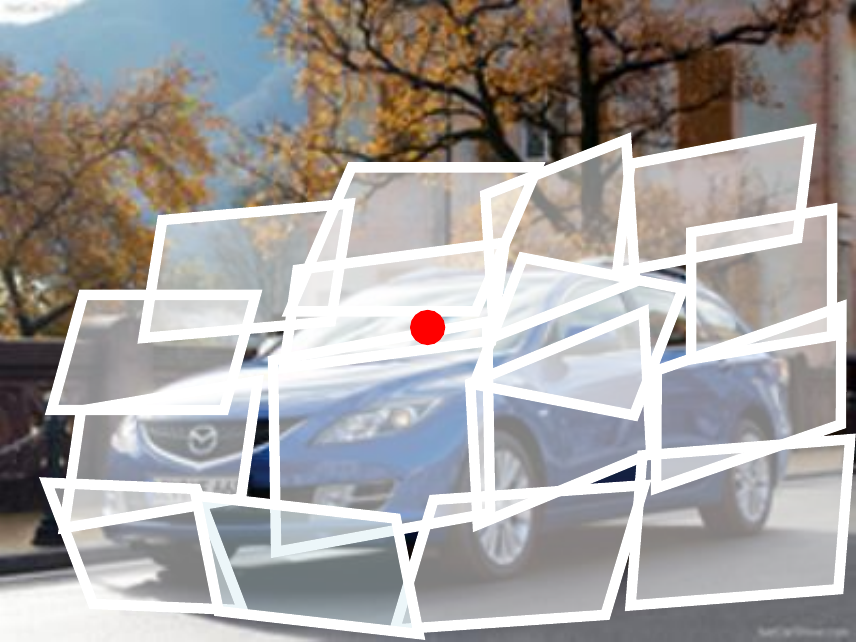}}\hfill
	\subfigure[(e)]
	{\includegraphics[width=0.164\linewidth]{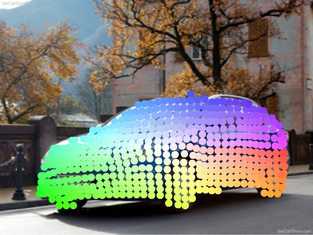}}\hfill
	\subfigure[(f)]
	{\includegraphics[width=0.164\linewidth]{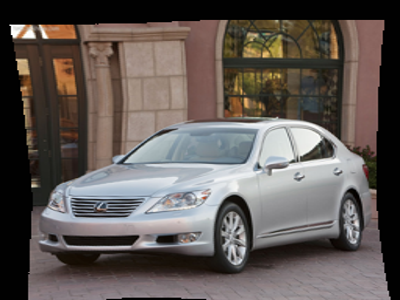}}\hfill
	\vspace{-10pt}	
	\caption{Visualization of pyramidal model in the PARN : (a) source and target images, estimated affine field at (b) level 1, (c) level 2, (d) level 3, (e) pixel-level, and (f) warped images. In each grid at each level, PARN estimates corresponding affine transformation field regularized with the estimated transformation field at previous level.}\label{img:1}\vspace{-10pt}
\end{figure}

In addition to the effort at measuring reliable matching evidences across images
under intra-class appearance variations, recent CNN-based approaches
have begun directly regressing geometric deformation fields through deep networks \cite{deephomographynetwork,Rocco17}.
As pioneering works, spatial transformer networks (STNs)
\cite{Jaderberg15} and its variant, inverse compositional spatial transformer networks (IC-STNs) \cite{lin2016inverse}, offer a way to deal with geometric variations
within CNNs. Rocco et al. \cite{Rocco17} and Schneider et al. \cite{schneider2017regnet} developed a CNN architecture for geometry-invariant matching that estimates
transformation parameters across semantically similar images and different modalities.
However, these methods assume the global transformation model, and
thus they cannot deal with spatially-varying geometric variations,
which frequently appear in dense semantic correspondence. More
recently, some methods such as universal correspondence network (UCN) \cite{ucn} and deformable convolutional networks (DCN) \cite{dcn} were proposed to
encode locally-varying geometric variations in CNNs, but they do not
have smoothness constraints with neighboring points, and cannot
guarantee reliable performance under relatively large
geometric variations. An additional challenge lies in the
lack of training data with ground-truth for semantic correspondence,
making the use of supervised training approaches difficult.

In this paper, we present a novel CNN architecture, called pyramidal
affine regression networks (PARN), that estimates locally-varying	
affine transformation fields across semantically similar images in a coarse-to-fine fashion, as shown in \figref{img:1}.
Inspired by pyramidal graph models \cite{Kim13,Hur15} that impose the hierarchical
smoothness constraint on labeling results, our approach first estimates a global affine
transformation over an entire image, and then progressively
increases the degree of freedom of the transformation in a form of quad-tree, 
finally producing pixel-wise continuous affine transformation
fields. The regression networks estimate residual affine transformations at each level
and these are composed to provide final affine transformation fields. To
overcome the limitations of insufficient training data for semantic
correspondence, we propose a novel weakly-supervised training scheme that generates progressive supervisions by leveraging the correspondence consistency. Our method works in an end-to-end manner, and does not require
quantizing the search space, different from conventional methods
\cite{Tola10,Hur15}. To the best of our knowledge, it is the first attempt to estimate the locally-varying affine transformation fields
through deep network in a coarse-to-fine manner. Experimental
results show that the PARN outperforms the latest methods for dense
semantic correspondence on several benchmarks including Taniai dataset \cite{Taniai16}, PF-PASCAL \cite{Ham17}, and Caltech-101 \cite{li06}. \vspace{-5pt}

\section{Related Works}\label{sec:2}

\subsubsection{Dense Semantic Correspondence}\label{sec:21}
Liu et al.~\cite{Liu11} pioneered the idea of dense correspondence across different scenes, and proposed SIFT Flow.
Inspired by this, Kim et al.~\cite{Kim13} proposed the deformable spatial pyramid (DSP) which performs multi-scale regularization within a hierarchical graph. More recently, Yang et al. \cite{yang2017object} proposed the object-aware hierarchical graph (OHG) to regulate matching consistency over whole objects. Among other methods are those that take an exemplar-LDA approach \cite{Bristow15}, employ joint image set alignment \cite{Zhou15}, or jointly solve for cosegmentation \cite{Taniai16}. As all of these techniques use handcrafted descriptors such as SIFT \cite{Lowe04} or DAISY \cite{Tola10}, they lack the robustness to deformations that is possible with deep CNNs.

Recently CNN-based descriptors have been used to establish dense semantic correspondences because of their high invariance to appearance variations. 
Zhou et al.~\cite{Zhou16} proposed a deep network that exploits cycle-consistency with a 3-D CAD model \cite{ShapeNet} as a supervisory signal. Choy et al.~\cite{ucn} proposed the universal correspondence network (UCN) based on fully convolutional feature learning. Novotny et al.~\cite{anchornet} proposed AnchorNet that learns geometry-sensitive features for semantic matching with weak image-level labels. Kim et al.~\cite{fcss} proposed the FCSS descriptor that formulates local self-similarity  within a fully convolutional network.
However, none of these methods is able to handle severe non-rigid geometric variations.
\vspace{-10pt}
\subsubsection{Transformation Invariance}\label{sec:22}
Several methods have aimed to alleviate geometric variations through extensions of SIFT Flow, including scale-less SIFT Flow (SLS) \cite{Hassner12}, scale-space SIFT Flow (SSF) \cite{Qiu14}, and generalized DSP \cite{Hur15}.
However, these techniques have a critical and practical limitation that their computational cost increases linearly with the search space size. HaCohen et al.~\cite{HaCohen2011} proposed 
in a non-rigid dense correspondence (NRDC) algorithm, but it employs weak matching evidence that cannot guarantee reliable performance. Geometric invariance to scale and rotation is provided by DAISY Filer Flow (DFF)~\cite{Yang14}, but its implicit smoothness constraint often induces mismatches. Recently, Ham et al.~\cite{Ham16} presented the Proposal Flow (PF) algorithm to estimate correspondences using object proposals. Han et al. \cite{han17} proposed SCNet to learn the similarity function and geometry kernel of PF algorithm within deep CNN.
While these aforementioned techniques provide some amount of geometric invariance, none of them can deal with
affine transformations across images, which frequently occur in dense semantic correspondence. 
More recently, Kim et al. \cite{dctm} proposed DCTM framework where dense affine transformation fields are inferred using a handcrafted energy function and optimization.

STNs \cite{Jaderberg15} offer a way to deal with geometric variations within CNNs
by warping features through a global parametric
transformation. Lin et al. \cite{lin2016inverse}
proposed IC-STNs that replaces the feature warping with transformation
parameter propagation. Rocco et al.~\cite{Rocco17} proposed a CNN architecture for estimating a geometric model such as an affine transformation for semantic correspondence estimation. 
However, it only estimates globally-varying geometric fields, and thus exhibits limited performance for dealing with locally-varying geometric deformations. Some methods such as UCN \cite{ucn} and DCN \cite{dcn} were proposed to
encode locally-varying geometric variations in CNNs, but they do not
have the smoothness constraints with neighboring points and cannot
guarantee reliable performance for images with relatively large
geometric variations \cite{dctm}.

\section{Method}\label{sec:3}
\subsection{Problem Formulation and Overview}\label{sec:31}
Given a pair of images $I$ and $I'$,
the objective of dense correspondence estimation is to establish a correspondence $i'$ for each pixel $i=[i_\mathbf{x},i_\mathbf{y}]$.
In this work,
we infer a field of affine transformations, each represented by a $2\times3$ matrix
\begin{equation}
	\mathbf{T}_i = \left[ {\begin{array}{*{20}{c}}
			{\mathbf{T}_{i,\mathbf{x}}}\\
			{\mathbf{T}_{i,\mathbf{y}}}
		\end{array}} \right]
\end{equation}
that maps pixel $i$ to ${i'} = \mathbf{T}_{i}\mathbf{i}$, where $\mathbf{i}$ is pixel $i$ represented in homogeneous coordinates such that $\mathbf{i}=[i,1]^T$.
	
Compared to the constrained geometric transformation model (i.e. only translational motion) commonly used in the
stereo matching or optical flow estimation, the affine
transformation fields can model the geometric variation in a more
principled manner. Estimating the pixel-wise affine transformation
fields, however, poses additional challenges due to its
infinite and continuous solution space. It is well-known in stereo matching
literatures that global approaches using the
smoothness constraint defined on the Markov random field (MRF)
\cite{Li15} tend to achieve higher accuracy on the labeling
optimization, compared to local approaches based on the
structure-aware cost aggregation \cite{hosni13}. However, such global approaches do
not scale very well to our problem in terms of computational
complexity, as the affine transformation is defined over the 6-D
continuous solution space. Additionally, it is not easy to guarantee
the convergence of affine transformation fields estimated through
the discrete labeling optimization due to extremely large label
spaces. Though randomized search and propagation strategy for
labeling optimization \cite{Li15,Lu13} 
\begin{figure}[t!]
	\centering
	\renewcommand{\thesubfigure}{}
	\subfigure[]
	{\includegraphics[width=1\linewidth]{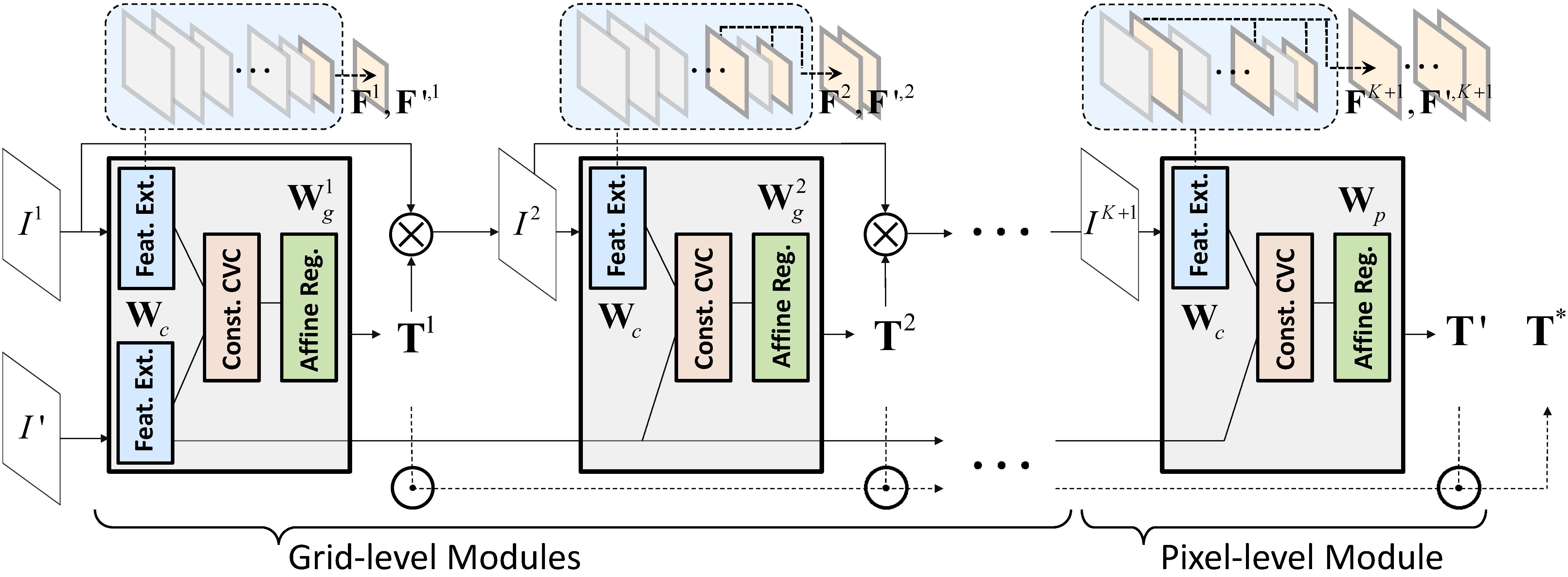}}\hfill	
	\vspace{-20pt}
	\caption{Network configuration of the PARN, which is defined on the pyramidal model and consists of several grid-level modules and a single pixel-level module. Each module is designed to mimic the standard matching process within a deep architecture, including feature extraction, cost volume construction, and regression.}\label{img:2}\vspace{-10pt}
\end{figure}
may help to improve
the convergence of labeling optimization on high-dimensional label
space, most approaches just consider relatively lower-dimensional
label space, e.g. 4-D label space consisting of translation,
rotation, and scale.
	
	Inspired by the pyramidal graph model \cite{Kim13,Hur15,Revaud15}
	and the parametric geometry regression networks \cite{deephomographynetwork,Rocco17}, we propose a novel
	deep architecture that estimates dense affine
	transformation fields in a coarse-to-fine manner. Our key
	observation is that affine transformation fields estimated at a
	coarse scale tend to be robust to geometric variations while the
	results at a fine scale preserve fine-grained details of objects
	better. While conventional approaches that employ the coarse-to-fine scheme in dense correspondence estimation \cite{Liu11,ranjan17} focus on \emph{image scales}, our approach exploits \emph{semantic scales} within the
	hierarchy of deep convolutional networks. Our method
	first estimates an image-level affine transformation using the
	deepest convolutional activations and then progressively localizes
	the affine transformation field additionally using the shallower convolutional
	activations in a quad-tree framework, producing the pixel-level
	affine transformation fields as the final labeling results.
	
	As shown in \figref{img:2}, our method is defined on the
	\emph{pyramidal model} (see \figref{img:1}) that consists of two kind of networks,
	several grid-level modules and a single pixel-level module, similar to
	\cite{Kim13,Hur15}. Each module within two networks is designed to
	mimic the standard matching process within a deep architecture \cite{Rocco17}:
	feature extraction, correlation volume construction, and regression.
	Concretely, when two images $I$ and $I'$ are given, convolutional features
	are first extracted as multi-level intermediate activations through
	the feature network (with ${{\bf{W}}_c}$) in order to provide fine-grained
	localization precision ability at each level while preserving
	robustness to deformations. Then, the correlation volume is
	constructed between these features at the cost volume construction layer of \figref{img:2}. Finally the affine transformation fields are inferred by passing the
	correlation volume to the regression network (with ${\bf{W}}^k_g$, ${\bf{W}}_p$
	of \figref{img:2}).
	This procedure is repeated for $K$ grid-level modules and a
	single pixel-level module.

	\subsection{Pyramidal Affine Regression Networks}\label{sec:32}
	Each module of our pyramidal model has three main components. 
	The first one extracts \textit{hierarchically} concatenated features from the input images and the second computes a cost volume within \textit{constrained} search windows.
	Lastly, from the third one, a \textit{locally-varying} affine field is densely estimated for all pixels.
	 \vspace{-10pt}
	 	
	\subsubsection{Feature Extraction}
	While conventional CNN-based descriptors have shown the excellent capabilities in handling intra-class appearnce variations \cite{imagenet,He16}, they have difficulites in yielding both semantic robustness and matching precision ability at the same time.
	To overcome this limitation, our networks are designed to leverage
	the inherent hierarchies of CNNs where multi-level intermediate
	convolutional features are extracted through a shared siamese network.
	We concatenate some of these convolutional feature maps such that
	\begin{equation}
		{\mathbf{F}^k = {\bigcup\nolimits_{n \in
					M(k)}} \mathcal{F} (I^{k};\mathbf{W}^n_c)}	
	\end{equation}
	where $\bigcup$ denotes the concatenation operator, $\mathbf{W}^n_c$ is the feature extraction network parameter until $n$-th convolutional layer and $M(k)$ is the sampled indices of convolutional layers at level $k$.
	This is illustrated by the upper of \figref{img:2}.

	Moreover, iteratively extracting the features along our pyramidal model provides evolving receptive fields which is a key ingradient for the geometric invariance \cite{Yang14,dctm}.
	By contrast, exitsting geometry regression networks \cite{Rocco17,deephomographynetwork} face a tradeoff between appearance invariance and localization precision due to the fixed receptive field of extracted features.
	Note that we obtained $I^k$ with the outputs from the previous level by warping $I^{k-1}$ with $\mathbf{T}^{k-1}$
	through bilinear samplers \cite{Jaderberg15} which facilitate an end-to-end learning framework.
	 \vspace{-10pt}
	 
	\subsubsection{Constrained Cost Volume Construction}
	To estimate geometry between image pairs $I^k$ and $I'$, the matching cost according to search spaces should
	be computed using extracted features ${\mathbf F}^k$ and
	${\mathbf F}'^{,k}$. Unlike conventional approaches that quantize
	search spaces for estimating depth \cite{taxonomy}, optical flow
	\cite{Butler12}, or similarity transformations
	\cite{Hur15}, quantizing the 6-D affine transformation
	defined over an infinite continuous solution space is
	computationally expensive and also degenerates the estimation
	accuracy. Instead, inspired by traditional robust geometry
	estimators such as RANSAC \cite{ransac} or Hough voting \cite{Lowe04}, we first
	construct the cost volume computed with respect to translational
	motion only, and then determine the affine transformation for each block
	by passing it through subsequent convolutional layers to reliably
	prune incorrect matches.
	
	\begin{figure}[t!]
		\centering
		\renewcommand{\thesubfigure}{}
		\subfigure[(a)]
		{\includegraphics[width=0.164\linewidth]{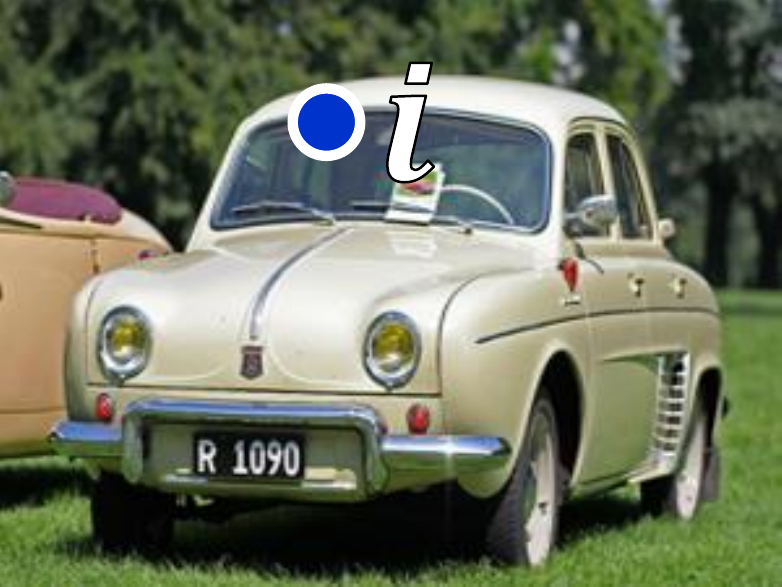}}\hfill
		\subfigure[(b)]    
		{\includegraphics[width=0.164\linewidth]{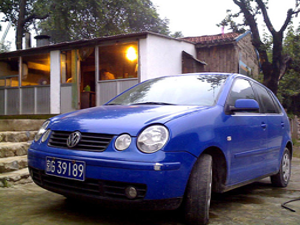}}\hfill
		\subfigure[(c)]  
		{\includegraphics[width=0.164\linewidth]{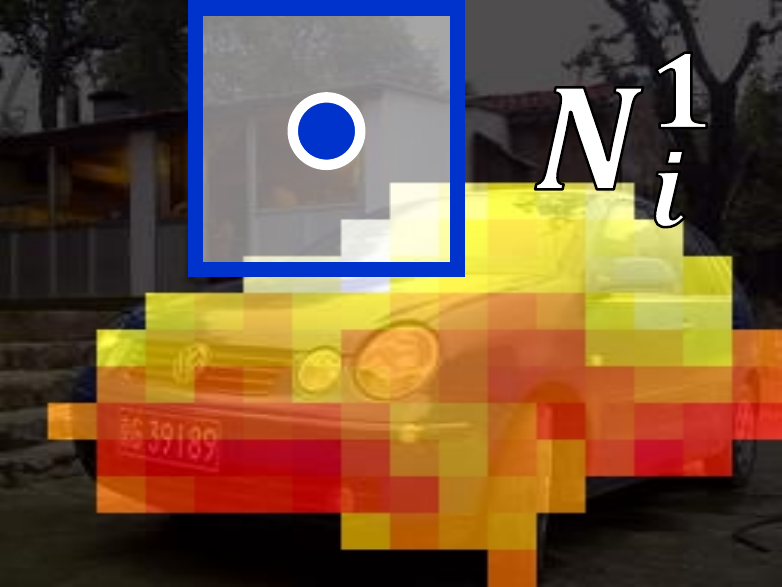}}\hfill
		\subfigure[(d)] 
		{\includegraphics[width=0.164\linewidth]{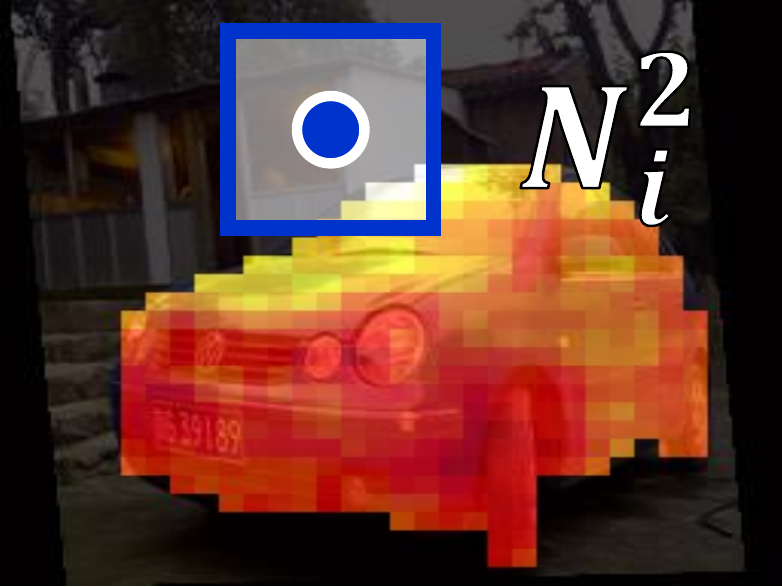}}\hfill
		\subfigure[(e)] 
		{\includegraphics[width=0.164\linewidth]{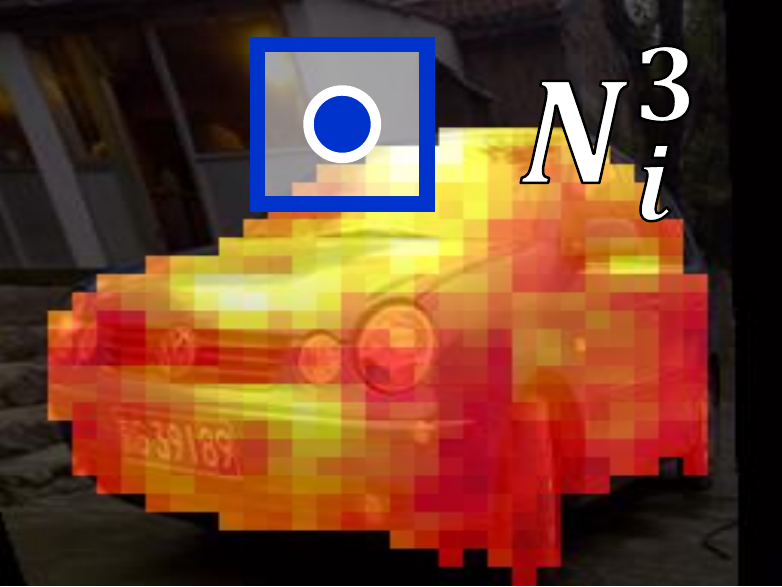}}\hfill
		\subfigure[(f)] 
		{\includegraphics[width=0.164\linewidth]{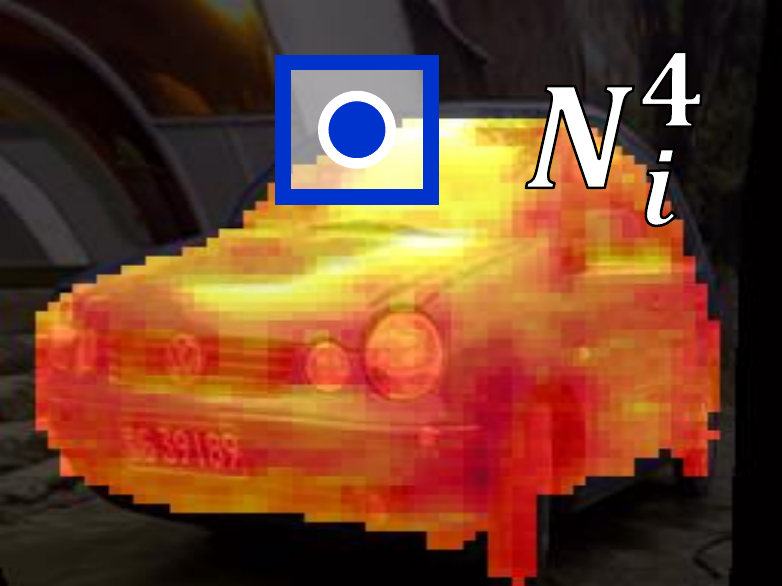}}\hfill
		\vspace{-10pt}
		\caption{Visualization of the constrained search window $N^{k}_i$: (a) source image and a reference pixel (blue colored). The matching costs are visualized as the heat maps for the reference pixel at (c) level 1, (d) level 2, (e) level 3, and (f) pixel-level.}
		\label{img:3}\vspace{-10pt}
	\end{figure}
	
	Concretly, the matching costs between extracted features ${\mathbf F}^k$,
	${\mathbf F}'^{,k}$ are computed within a search window as a rectified cosine similarity, such that
	\begin{equation}
		\mathbf{C}^{k}(i,j) = \max (0,\mathbf{F}'^{,k}(i) \cdot \mathbf{F}^k(j)), \quad\text{where}\quad j\in N^{k}_i.
	\end{equation}
	A constrained search window $N^{k}_i$ is centered at pixel $i$ with the radius $r(k)$ as examplified in \figref{img:3}.
	In our pyramidal model, a
	relatively large radius is used at coarser levels to estimate a
	rough yet reliable affine transform as a guidance at subsequent
	finer levels. The radius becomes smaller as the level goes
	deeper where the regression network is likely to avoid local minima
	thanks to the guidance of affine transformation fields estimated on
	the previous level.
	Thus only reliable matching candidates are provided as an input to the following regression network where even fine-scaled geometric transformations can be estimated at deeper level.
	The constructed cost volume can be further utilized for generating the supervisions with correspondence consistency check as described in \secref{sec:32}.

	\vspace{-10pt}
	
	\subsubsection{Grid-level Regression}
	The constrained cost volume $\mathbf{C}^k$ is passed through successive CNNs and bilinear upsampling layer to estimate the affine transformation field
    such that $\mathbf{T}^k = \mathcal{F}(\mathbf{C}^k;\mathbf{W}^k_g)$,
    where $\mathbf{W}^k_g$ is the grid-level regression network parameter at the level $k$.
	Since each level in the pyramid has a simplified task (it only has to estimate residual transformation field), the regression networks can be simple to have 3-6 convolutional layers.
	
	Within the hierarchy of the pyramidal model, our first starts
	to estimate the transformation from an entire image and then
	progressively increase the degree of freedom of the transformation
	by dividing each grid into four rectangular grids, yielding
	$2^{k-1}\times2^{k-1}$ grid of affine fields at level $k$. 
	However, the estimated coarse affine field has the discontinuities between nearby affine fields occuring blocky artifacts around grid boundaries as shown in (d) and (f) of \figref{img:6}.
	To alleviate this, a bilinear upsampler \cite{Jaderberg15} is applied at the end of successive CNNs, upsampling a coarse grid-wise affine field to the original resolution of the input image $I$.
	This simple strategy regularizes the affine field to be smooth, suppressing the artifacts considerably as examplified in \figref{img:6}.

	Note that the composition of the estimated affine fields from level 1 to $k$ can be computed as multiplications of augmented matrix in homogeneous coordinates such that 
	\begin{equation}
	\mathbf{M}(\mathbf{T}_i^{[1,k]})=\prod\nolimits_{n \in \{1,...,k\}} \mathbf{M}(\mathbf{T}_i^n)
	\end{equation}
	where $\mathbf{M}(\mathbf{T})$ represents $\mathbf{T}$ in homogeneous coordinates as $[\mathbf{T};[0,0,1]]$.
    \vspace{-10pt}
	
	\subsubsection{Pixel-level Regression}
	To improve the matching ability localizing fine-grained object
	boundaries, we additionally formulate a pixel-level module.
	Similar to the grid-level modules, it also consists of
	feature extraction, constrained cost volume construction, and
	regression network.
	The main difference is that an encoder-decoder architecture is employed for the regression network,
	which has been adopted in many
	pixel-level prediction tasks such as disparity estimation \cite{godard17}, optical
	flow \cite{fischer15}, or semantic segmentation \cite{noh15}.
	Taking a warped image $I^{K+1}$ as an input, a constrained cost volume $\mathbf{C}^{K+1}$ is computed and 
	the pixel-level affine field is regressed through the encoder-decoder network such that $\mathbf{T}'=\mathcal{F}
	(\mathbf{C}^{K+1};\mathbf{W}_p)$, where $\mathbf{W}_p$ is the pixel-level regression network parameter.
	The final affine transformation field between source and target image can be computed as ${\bf{M}}({\bf{T}}_i^*) = {\bf{M}}({\bf{T}}_i^{[1,K]})\cdot{\bf{M}}({{\bf{T'}}_i})$.
	\begin{figure}[t!]
		\centering
		\renewcommand{\thesubfigure}{}
		\subfigure[]
		{\includegraphics[width=0.8\linewidth]{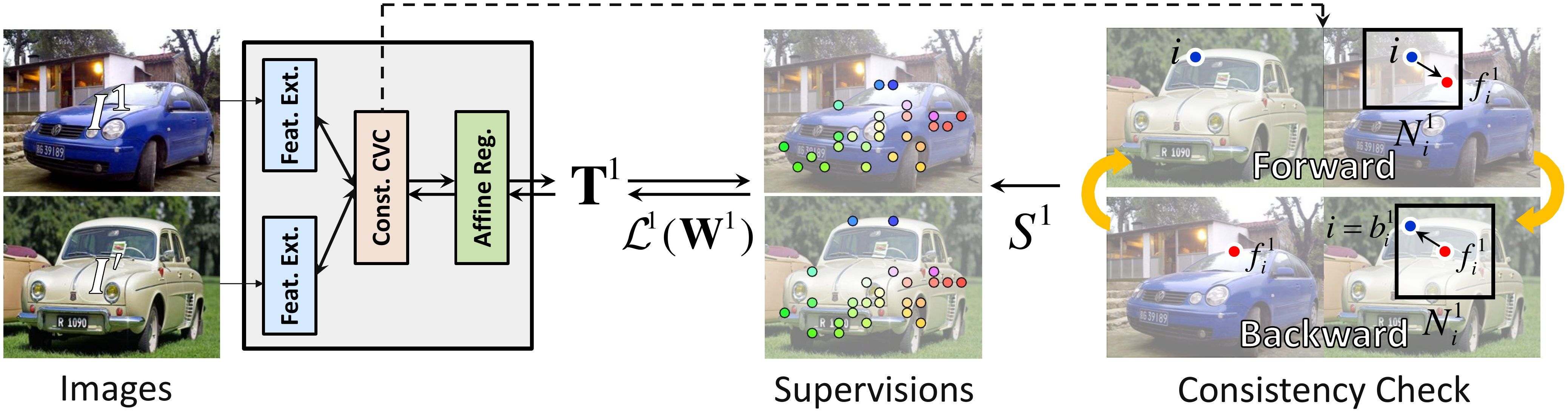}}
		\vspace{-20pt}
		\caption{Training the grid-level module at level $1$. By using the correspondence consistency, tentative sparse correspondences are determined and used to train the network.}
		\label{img:4}
	\end{figure}
	\begin{figure}[t!]
		\centering
		\renewcommand{\thesubfigure}{}
		\subfigure[]
		{\includegraphics[width=0.164\linewidth]{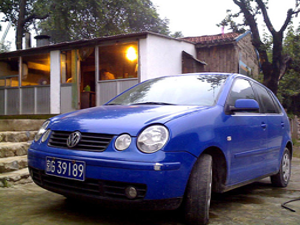}}\hfill
		\subfigure[]
		{\includegraphics[width=0.164\linewidth]{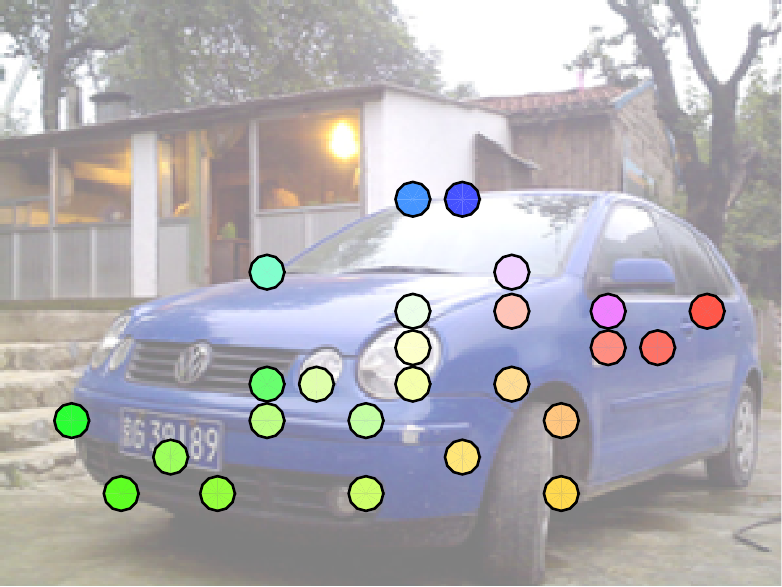}}\hfill
		\subfigure[]
		{\includegraphics[width=0.164\linewidth]{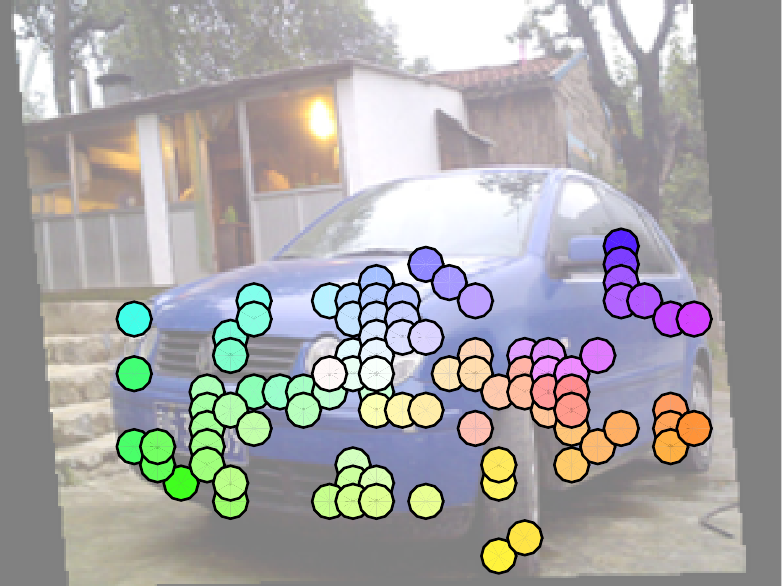}}\hfill
		\subfigure[]
		{\includegraphics[width=0.164\linewidth]{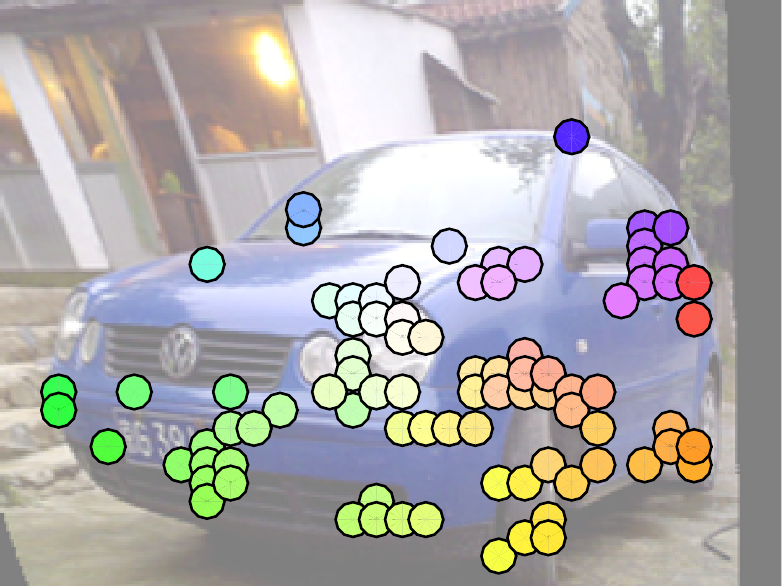}}\hfill
		\subfigure[]
		{\includegraphics[width=0.164\linewidth]{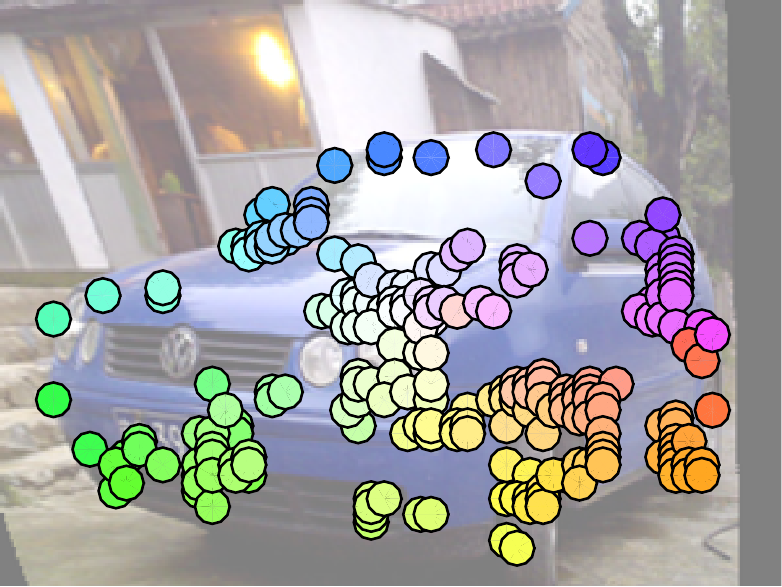}}\hfill
		\subfigure[]
		{\includegraphics[width=0.164\linewidth]{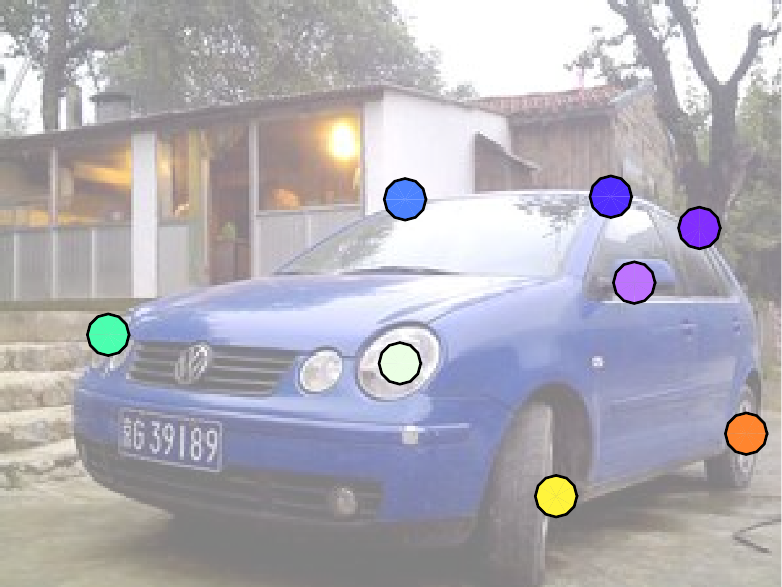}}\hfill
		\vspace{-21pt}
		\subfigure[(a)]
		{\includegraphics[width=0.164\linewidth]{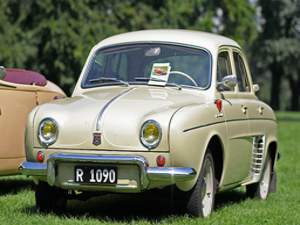}}\hfill
		\subfigure[(b)]
		{\includegraphics[width=0.164\linewidth]{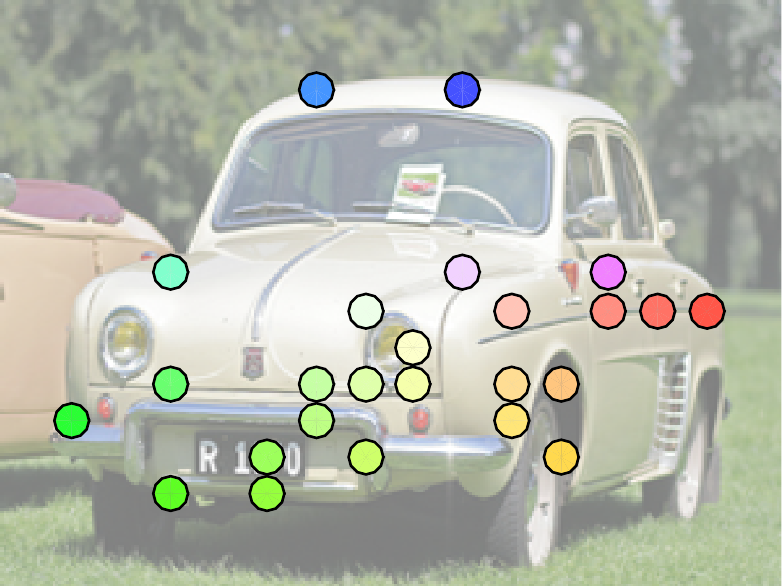}}\hfill
		\subfigure[(c)]
		{\includegraphics[width=0.164\linewidth]{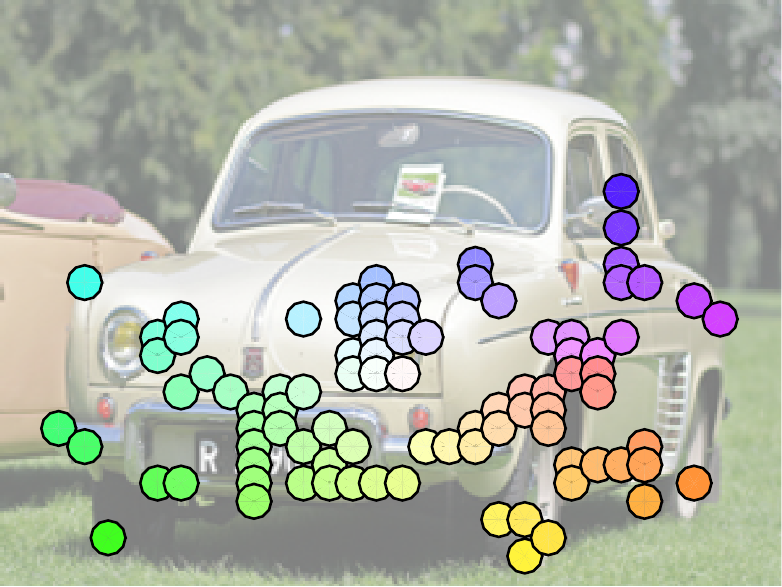}}\hfill
		\subfigure[(d)]
		{\includegraphics[width=0.164\linewidth]{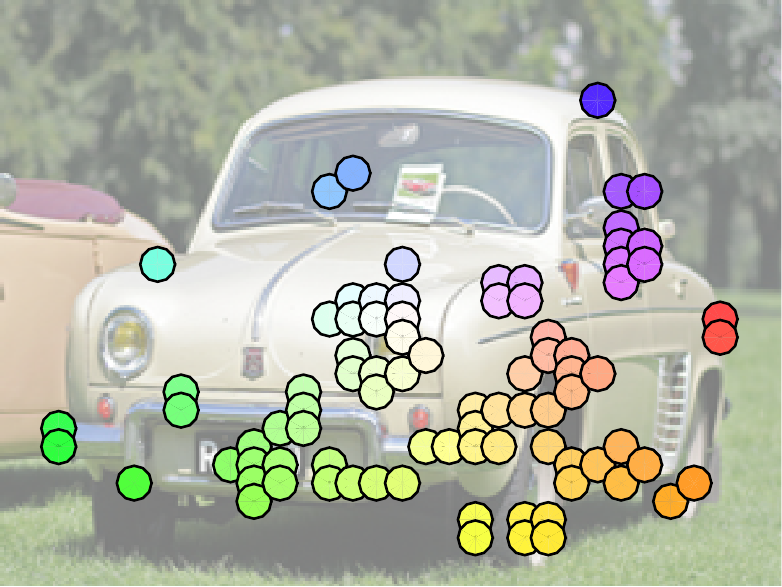}}\hfill
		\subfigure[(e)]
		{\includegraphics[width=0.164\linewidth]{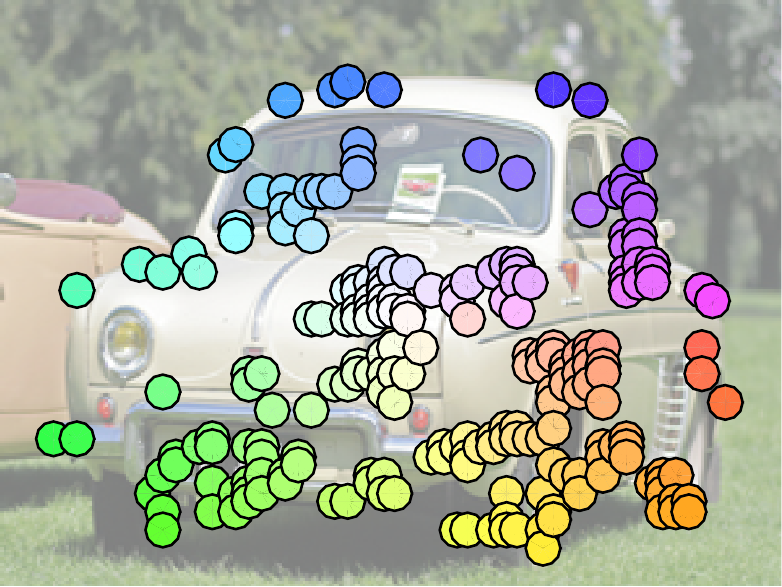}}\hfill
		\subfigure[(f)]
		{\includegraphics[width=0.164\linewidth]{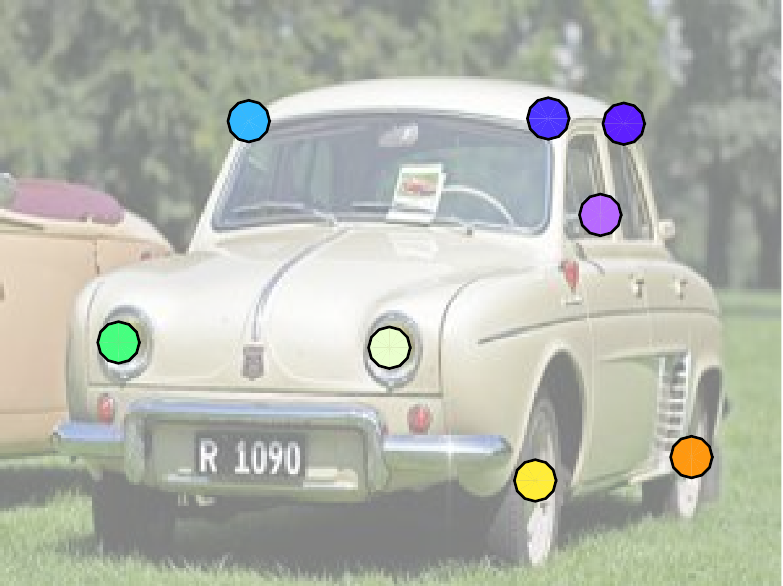}}\hfill
		\vspace{-10pt}
		\caption{Visualization of the generated supervisions at each level: (a) source and target images, (b) level 1, (c) level 2, (d) level 3, (e) pixel level, (f) GT keypoints. The tentative positive samples are color-coded. (Best viewed in color.)}\label{img:5}\vspace{-10pt}
	\end{figure}
	\begin{table}[t!]
		\begin{center}
			\begin{tabularx}{\linewidth}{p{3.0mm} p{5.0mm}| p{2.0mm}| p{2.0mm}| p{160mm}}
				\hlinewd{0.8pt}
				\multicolumn{5}{ p{164mm} }{{\bf Algorithm 1}: Pyramidal Affine Regression Network}\\
				\hlinewd{0.8pt}
				\multicolumn{5}{ p{164mm} }{{\bf Input}: images $I$, $I'$}\\
				\multicolumn{5}{ p{164mm} }{{\bf Output}: network parameters ${\mathbf W}_{c}$, ${\mathbf W}_{g}$, $\mathbf{W}_p$, affine fields $\mathbf{T}^*$}\\
				$\mathbf{1:}$&\multicolumn{4}{ p{163mm} }{\quad Compute convolutional activations of target image $I'$}\\
				~&\multicolumn{4}{ p{163mm} }{\quad {\bf for} $k=1:K$ {\bf do }}\\
				$\mathbf{2:}$&~&\multicolumn{3}{ p{158mm} }{\quad Compute image $I^{k}$ by warping $I^{k-1}$ with $\mathbf{T}^{k-1}$ when $k>1$}\\
				$\mathbf{3:}$&~&\multicolumn{3}{ p{158mm} }{\quad \textbf{[Only when training]} : Initialize affine fields as $\mathbf{T}^k_i=[\mathbf{I}_{2\times2},\mathbf{0}_{2\times1}]$}\\
				~&~&\multicolumn{3}{ p{162mm} }{\quad {\bf $/*$ \emph{Feature Extraction} $*/$}}\\
				$\mathbf{4:}$&~&\multicolumn{3}{ p{158mm} }{\quad Compute convolutional activations of $I^k$ and extract features $\mathbf{F}^{k}$, $\mathbf{F}'^{,k}$}\\
				~&~&\multicolumn{3}{ p{162mm} }{\quad {\bf $/*$ \emph{Constrained Correlation Volume} $*/$}}\\
				$\mathbf{5:}$&~&\multicolumn{3}{ p{158mm} }{\quad Construct the constrained cost volume $\mathbf{C}^{k}$ with radius $r(k)$}\\
				$\mathbf{6:}$&~&\multicolumn{3}{ p{158mm} }{\quad \textbf{[Only when training]} : Generate supervisions $S^k$ and train the network}\\
				~&~&\multicolumn{3}{ p{162mm} }{\quad {\bf $/*$ \emph{Affine Transformation Field Regression} $*/$}}\\				
				$\mathbf{7:}$&~&\multicolumn{3}{ p{158mm} }{\quad \textbf{[Only when testing]} : Estimate affine fields ${{\mathbf{T}^k}} = {\mathcal F}(\mathbf{C}^{k};{\mathbf W}^{k}_{g})$}\\  
				~&\multicolumn{4}{ p{163mm} }{\quad {\bf end for }}\\
				$\mathbf{8:}$&\multicolumn{4}{ p{163mm} }{\quad Estimate pixel-level affine fields $\mathbf{T}'=\mathcal{F}(\mathbf{C}^{K+1};\mathbf{W}_p)$}\\
				$\mathbf{9:}$&\multicolumn{4}{ p{163mm} }{\quad Compute final affine fields $\mathbf{M}({\mathbf{T}^*_i})=\prod_{n \in \{1,...,K\}} \mathbf{M}(\mathbf{T}^n) \cdot \mathbf{M}(\mathbf{T}'_i)$}\\   
				\hlinewd{0.8pt}
			\end{tabularx}
		\end{center}\label{alg:1}\vspace{-20pt}
	\end{table}
	\subsection{Training}\label{sec:32}
	\subsubsection{Generating Progressive Supervisions}
	A major challenge of semantic correspondence with CNNs is the lack
	of ground-truth correspondence maps for training data. A possible
	approach is to synthetically generate a set of image pairs
	transformed by applying random transformation fields to make the
	pseudo ground-truth \cite{deephomographynetwork,Rocco17}, but this approach cannot reflect the \textit{realistic}
	appearance variations and geometric transformations well.
	
	Instead of using synthetically deformed imagery, we propose to generate supervisions
	directly from the \textit{semantically related} image pairs as shown in \figref{img:4},
    where the correspondence consistency check [35, 48] is applied to the constructed cost volume of each level.
	Intuitively, the correspondence relation from a source image to a target image should be consistent with that from the target image to the source image.
	Given the constrained cost volume ${\mathbf C}^k$, the best match $f_i^k$ is computed by searching the maximum score for each point $i$,
	$f_i^k = \mathop{\mathrm{argmax}}\nolimits_{j} {\mathbf C}^k(i,j)$.
	We also compute the backward best match $b_i^k$ for $f_i^k$ such that
	$b_i^k = \mathop{\mathrm{argmax}}\nolimits_{m} {\mathbf C}^k(m,f_i)$ to identify that the best match $f_i^k$ is consistent or not.
	By running this consistency check along our pyramidal model, we actively
	collect the tentative positive samples at each level such that $S^k=\{i|i=b_i^k, i\in \Omega\}$.
	We found that the generated supervisions are qualitatively and quantitatively superior to the sparse ground-truth keypoints as examplified in \figref{img:5}.
		
	For the accuracy of supervisions, we limit the
	correspondence candidate regions using object location priors
	such as bounding boxes or masks containing the target object to be matched,
	which are provided in most benchmarks \cite{li06,Everingham10,Chen14}.
	Note that our approach is conceptually similar to \cite{fcss},
	but we generate the supervisions from the constrained cost volume in a hiearchical manner so that the false positive samples are avoided which is critical to train the geometry regression network.
	
	\subsubsection{Loss Function}
	To train the module at level $k$, the loss function is
	defined as a distance between the flows at the positive samples and the flow fields computed by applying estimated
	affine transformation field such that
	\begin{equation}\label{equ:k-nn}
		\mathcal{L}^k(\mathbf{W}^k) = \frac{1}{N}\sum\nolimits_{i\in S^k}{\| \mathbf{T}^{k}_{i}\mathbf{i} - (i-f^k_i) \|^2},
	\end{equation}
	where $\mathbf{W}^k$ is the parameters of feature extraction network and regression network at level $k$
	and $N$ is the number of training samples.
	Algorithm 1 provides an overall summary of PARN.
	
	\section{Experimental Results}
	\subsection{Experimental Settings}
	For feature extraction networks in each regression module, we used the ImageNet pretrained VGGNet-16 \cite{Simonyan15} and ResNet-101 \cite{He16} with their network parameters.
	For the grid-level regressions, we used three grid-level modules ($K=3$), followed by a single pixel-level module.
	For $M(k)$ in the feature extraction step, we sampled convolutional activations after intermediate pooling layers such as `conv5-3', `conv4-3', and `conv3-3'. The radius of search space $r(k)$ is set to the ratio of the whole search space, and decreases as the level goes deeper such that $\left\lbrace 1/10, 1/10, 1/15, 1/15\right\rbrace$.
	\begin{figure}[t!]
		\centering
		\renewcommand{\thesubfigure}{}
		\subfigure[]
		{\includegraphics[width=0.123\linewidth]{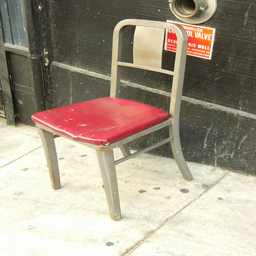}}\hfill
		\subfigure[]
		{\includegraphics[width=0.123\linewidth]{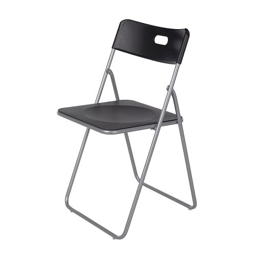}}\hfill
		\subfigure[]
		{\includegraphics[width=0.123\linewidth]{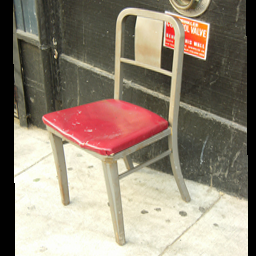}}\hfill
		\subfigure[]
		{\includegraphics[width=0.123\linewidth]{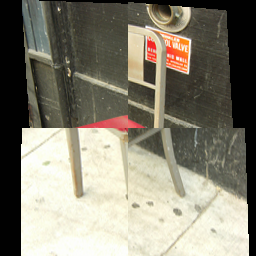}}\hfill
		\subfigure[]
		{\includegraphics[width=0.123\linewidth]{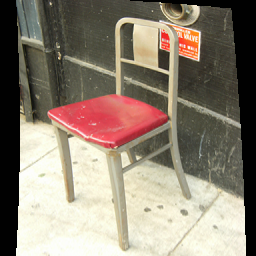}}\hfill
		\subfigure[]
		{\includegraphics[width=0.123\linewidth]{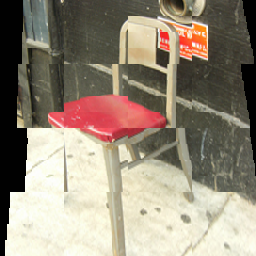}}\hfill
		\subfigure[]
		{\includegraphics[width=0.123\linewidth]{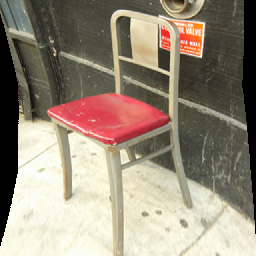}}\hfill
		\subfigure[]
		{\includegraphics[width=0.123\linewidth]{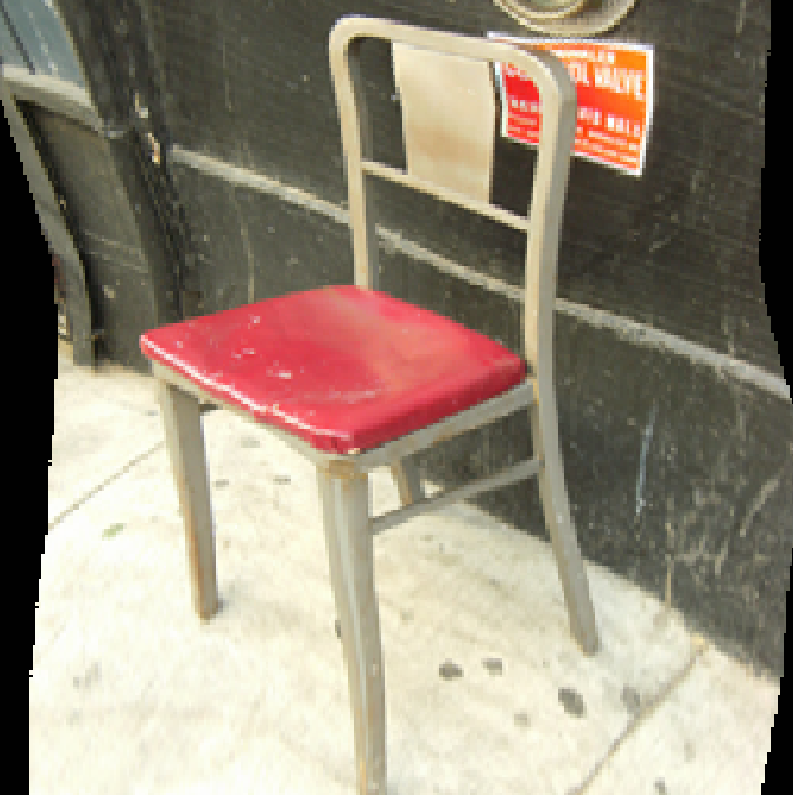}}\hfill
		\vspace{-21pt}
		\subfigure[(a)]
		{\includegraphics[width=0.123\linewidth]{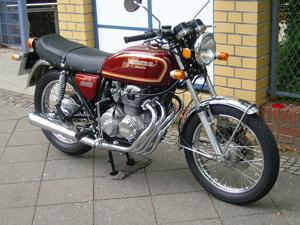}}\hfill
		\subfigure[(b)]
		{\includegraphics[width=0.123\linewidth]{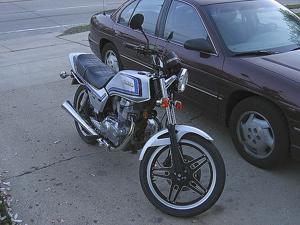}}\hfill
		\subfigure[(c)]
		{\includegraphics[width=0.123\linewidth]{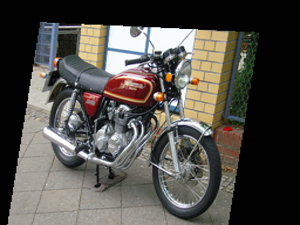}}\hfill
		\subfigure[(d)]
		{\includegraphics[width=0.123\linewidth]{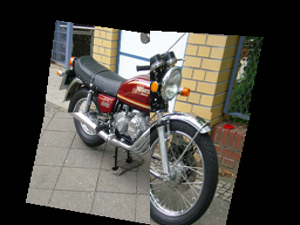}}\hfill
		\subfigure[(e)]
		{\includegraphics[width=0.123\linewidth]{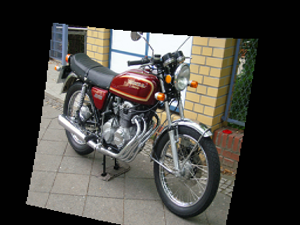}}\hfill
		\subfigure[(f)]
		{\includegraphics[width=0.123\linewidth]{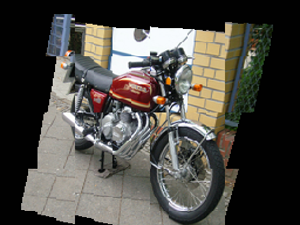}}\hfill
		\subfigure[(g)]
		{\includegraphics[width=0.123\linewidth]{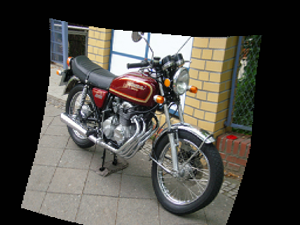}}\hfill
		\subfigure[(h)]
		{\includegraphics[width=0.123\linewidth]{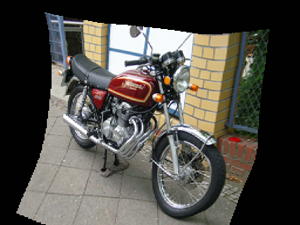}}\hfill
		\vspace{-10pt}
		\caption{Qualitative results of the PARN at each level: (a) source image, (b) target image, warping result at (c) level 1, (d) level 2 without upsampling layer, (e) level 2, (f) level 3 without upsampling layer, (g) level 3, and (h) pixel-level.}\label{img:6}\vspace{-10pt}
	\end{figure}
	
	In the following, we comprehensively evaluated PARN through comparisons to state-of-the-art dense semantic correspondences, including SIFT Flow \cite{Lowe04}, DSP \cite{Kim13}, and OHG \cite{yang2017object}.
	Furthermore, geometric-invariant methods including PF \cite{Ham16}, SCNet \cite{han17}, CNNGM \cite{Rocco17}, DCTM \cite{dctm}.
	The performance was measured on Taniai benchmark \cite{Taniai16}, PF-PASCAL dataset \cite{Ham17}, and Caltech-101 \cite{li06}. \vspace{-5pt}
	
	\subsection{Training Details}
	For training, we used the PF-PASCAL dataset \cite{Ham17} that consists of
	1,351 image pairs selected from PASCAL-berkely keypoint annotations
	of 20 object classes. We did not use the ground-truth keypoints at all to learn the network, but we utilized the
	masks for the accuracy of generated supervisions. We used 800 pairs as a training data, and further divide
	the rest of PF-PASCAL data into 200 validation pairs and 350 testing pairs.
	Additionally, we synthetically augment the training pair 10
	times by applying randomly generated geometric
	transformations including horizontal flipping \cite{Rocco17}.
	To generate the most accurate supervisions in the first level, we
	additionally apply M-estimator SAmple and Consensus (MSAC) \cite{msac} to
	build the initial supervisions $\mathbf{T}^0$ and restrict the
	search space with the estimated transformation.
	We sequentially trained the regression modules for 120k
	iterations each with a batch size of 16 and further finetune all the regression networks in an end-to-end manner \cite{lin2016inverse}. 
	The more details of experimental settings and training are provided in the supplemental material.
	
	\begin{figure}[t!]
		\centering
		\renewcommand{\thesubfigure}{}
		\subfigure[(a) FG3DCar]
		{\includegraphics[width=0.250\linewidth]{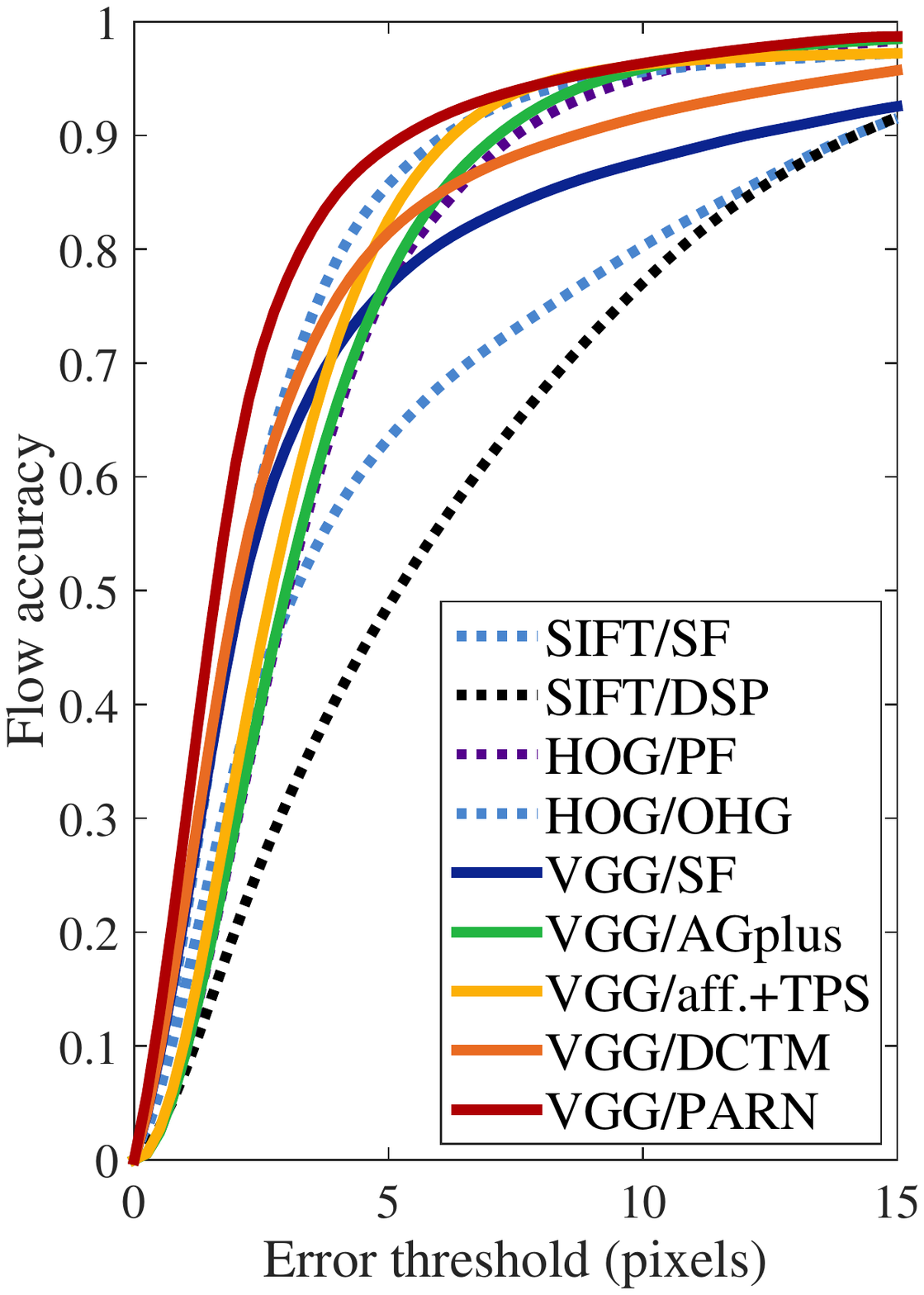}}\hfill
		\subfigure[(b) JODS]
		{\includegraphics[width=0.250\linewidth]{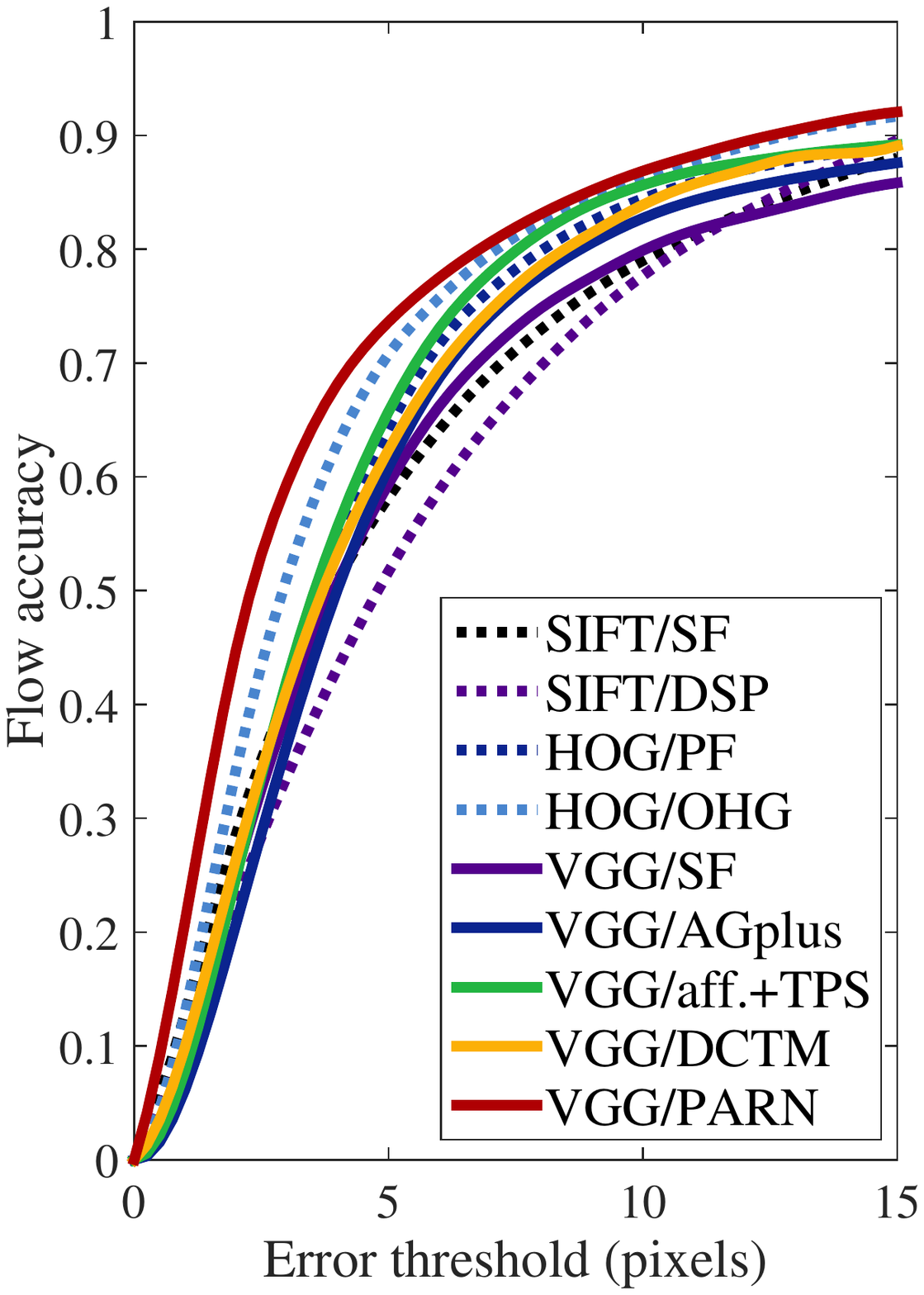}}\hfill
		\subfigure[(c) PASCAL]   
		{\includegraphics[width=0.250\linewidth]{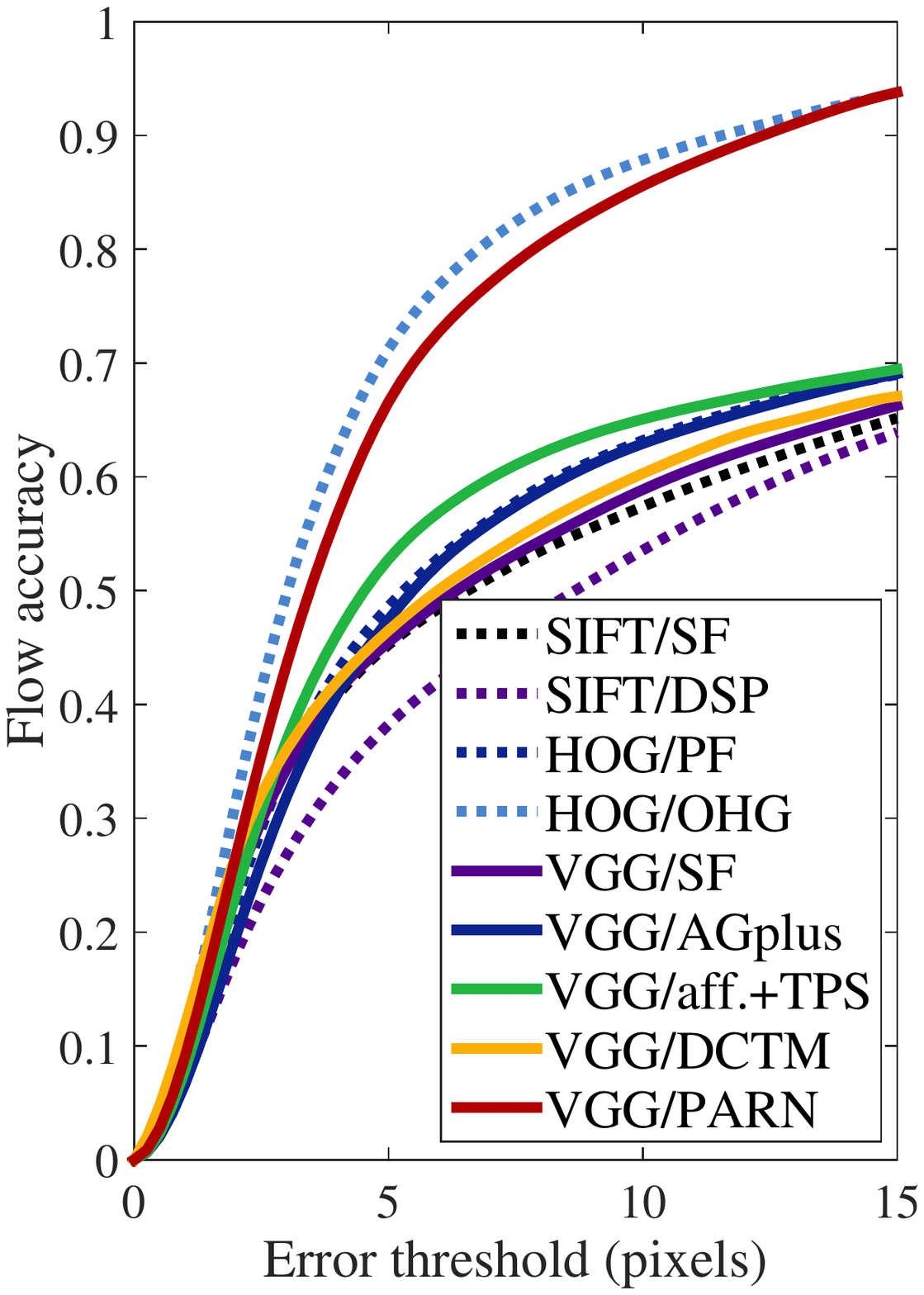}}\hfill
		\subfigure[(d) Average]
		{\includegraphics[width=0.250\linewidth]{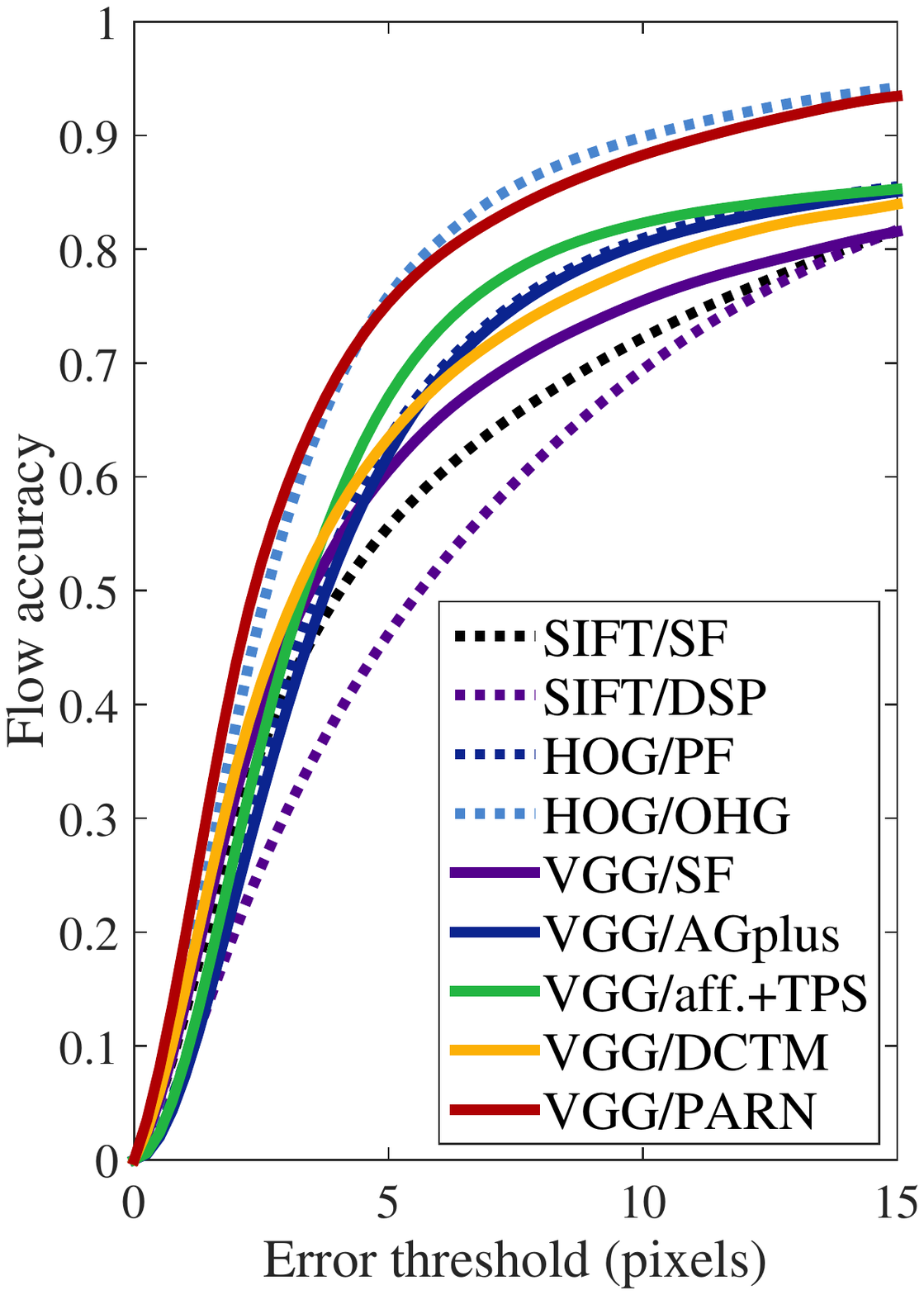}}\hfill
		\vspace{-10pt}
		\caption{Average matching accuracy with respect to endpoint error threshold on the Taniai benchmark \cite{Taniai16}: (from left to right) FG3DCar, JODS, PASCAL, and average.}\label{img:7}\vspace{-10pt}
	\end{figure}
	\begin{figure}[t!]
		\centering
		\renewcommand{\thesubfigure}{}
		\subfigure[]
		{\includegraphics[width=0.164\linewidth]{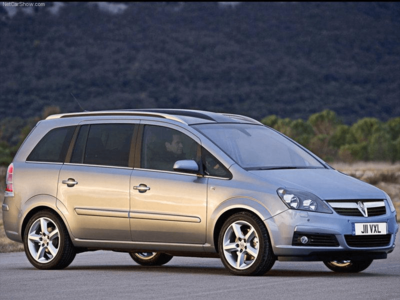}}\hfill
		\subfigure[]
		{\includegraphics[width=0.164\linewidth]{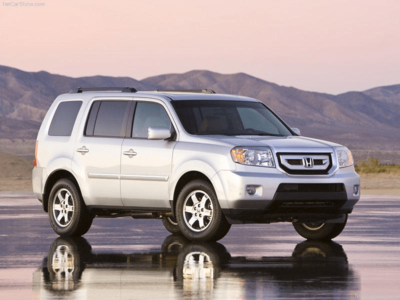}}\hfill
		\subfigure[]
		{\includegraphics[width=0.164\linewidth]{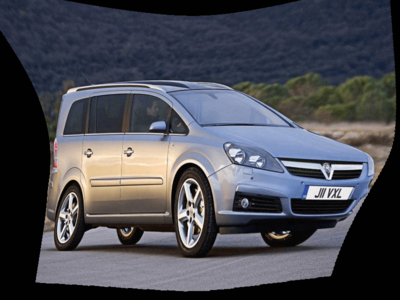}}\hfill
		\subfigure[]
		{\includegraphics[width=0.164\linewidth]{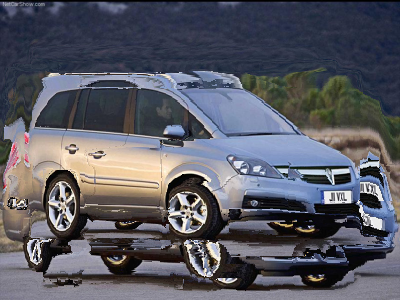}}\hfill
		\subfigure[]
		{\includegraphics[width=0.164\linewidth]{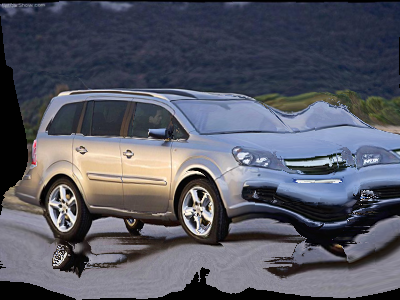}}\hfill
		\subfigure[]
		{\includegraphics[width=0.164\linewidth]{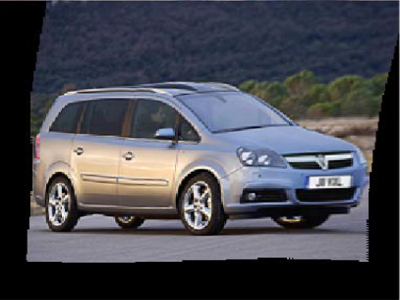}}\hfill
		\vspace{-21pt}
		\subfigure[(a)]
		{\includegraphics[width=0.164\linewidth]{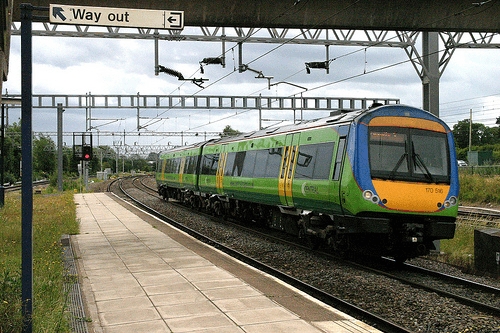}}\hfill
		\subfigure[(b)]
		{\includegraphics[width=0.164\linewidth]{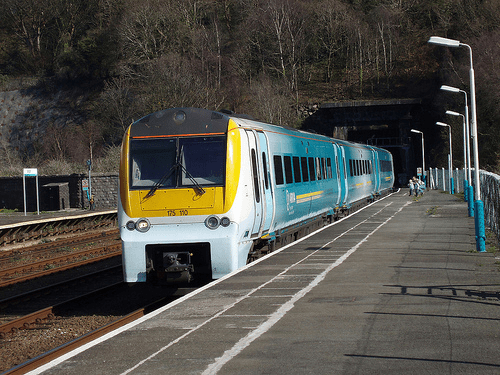}}\hfill
		\subfigure[(c)]
		{\includegraphics[width=0.164\linewidth]{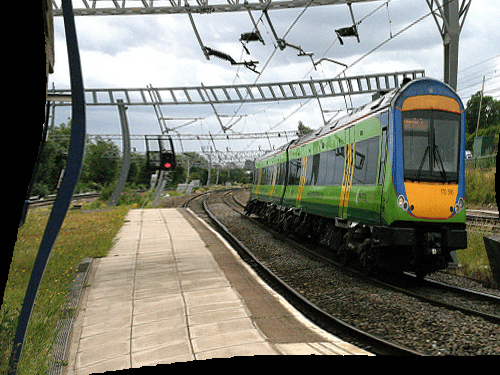}}\hfill
		\subfigure[(d)]
		{\includegraphics[width=0.164\linewidth]{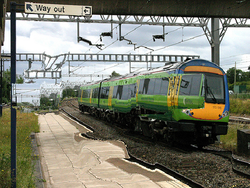}}\hfill
		\subfigure[(e)]
		{\includegraphics[width=0.164\linewidth]{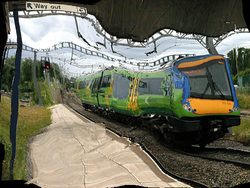}}\hfill
		\subfigure[(f)]
		{\includegraphics[width=0.164\linewidth]{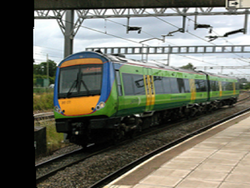}}\hfill
		\vspace{-10pt}
		\caption{Qualitative results on the Taniai benchmark
			\cite{Taniai16}: (a) source image, (b) target image, (c) CNNGM-Aff.TPS \cite{Rocco17}, (d) SCNet-AG+ \cite{han17},
			(e) DCTM \cite{dctm}, (f) PARN.
			The source images were warped to the target images using correspondences.}\label{img:8}
	\end{figure}			
	
	\begin{table}[!t]
		\centering
		\begin{tabular}{ >{\raggedright}m{0.15\linewidth} >{\centering}m{0.15\linewidth} >{\centering}m{0.15\linewidth}
				>{\centering}m{0.10\linewidth} >{\centering}m{0.10\linewidth}
				>{\centering}m{0.10\linewidth} >{\centering}m{0.10\linewidth}}
			\hlinewd{1.0pt}
			Methods &Descriptor &Matching &FG3D. &JODS &PASC. &Avg.\tabularnewline
			\hline
			\hline
			SF \cite{Liu11} &SIFT &SF &0.632 &0.509 &0.360 &0.500 \tabularnewline
			DSP \cite{Kim13} &SIFT &DSP &0.487 &0.465 &0.382 &0.445\tabularnewline
			PF \cite{Ham16} &HOG &LOM &0.786 &0.653 &0.531 &0.657 \tabularnewline
			OHG \cite{yang2017object} &HOG &OHG &0.875 &0.708 &\textbf{0.729} &0.771 \tabularnewline
			\hline
			\hline
			\multirow{3}{*}{SCNet \cite{han17}} &\multirow{3}{*}{VGG-16} &A &0.774 &0.574 &0.476 &0.608 \tabularnewline
			& &AG &0.764 &0.600 &0.463 &0.609 \tabularnewline
			& &AGplus &0.776 &0.608 &0.474 &0.619 \tabularnewline
			\hline
			\multirow{2}{*}{CNNGM \cite{Rocco17}} &\multirow{2}{*}{VGG-16} &Aff. &0.771 &0.662 &0.501 &0.644 \tabularnewline
			& &Aff.+TPS &0.835 &0.656 &0.527 &0.672 \tabularnewline
			\hline
			\multirow{2}{*}{DCTM \cite{dctm}} &VGG-16 &\multirow{2}{*}{DCTM} &0.790 &0.611 &0.528 &0.630 \tabularnewline
			&Affine-FCSS & &0.891 &0.721 &0.610 &0.740 \tabularnewline
			\hline
			\multirow{4}{*}{Baseline} &\multirow{4}{*}{VGG-16} &SF &0.756 &0.490 &0.360 &0.535 \tabularnewline
			& &PARN-Lv1 &0.783 &0.668 &0.641 &0.697 \tabularnewline
			& &PARN-Lv2 &0.837 &0.689 &0.656 &0.739 \tabularnewline
			& &PARN-Lv3 &0.869 &0.707 &0.681 &0.752 \tabularnewline
			\hline
			\multirow{2}{*}{Proposed} &VGG-16 &\multirow{2}{*}{PARN} &0.876 &0.716 &0.688 &0.760 \tabularnewline
			&ResNet-101 & &\textbf{0.895} &\textbf{0.759} &0.712 &\textbf{0.788} \tabularnewline
			\hlinewd{1.0pt}
		\end{tabular}\vspace{+6pt}
		\caption{Matching accuracy compared to state-of-the-art correspondence techniques on the Taniai benchmark \cite{Taniai16}.}\label{tab:1}\vspace{-20pt}
	\end{table}
	\subsection{Ablation Study}
	To validate the components within PARN,
	we additionally evaluated it at each level such as `PARN-Lv1', `PARN-Lv2', and `PARN-Lv3' as shown in \figref{img:6} and \tabref{tab:1}. For quantitative evaluations, we used the matching accuracy on the Taniai benchmark \cite{Taniai16}, which is described in details in the following section. As expected, even though the global transformation was estimated roughly well in the coarest level (i.e. level 1), the fine-grained matching details cannot be achieved reliably, thus showing the limited performance. However, as the levels go deeper, the localization ability has been improved while maintaining globally estimated transformations.
	The performance of the backbone network was also evaluated with a standard SIFT flow optimization \cite{Liu11}.
	Note that the evaluation of the pixel-level module only in our networks is impracticable, since it requires a pixel-level supervision that
	does not exist in the current public datasets for semantic correspondence.
	
	\subsection{Results}\label{sec:43}
	See the supplemental material for more qualitative results.
	\subsubsection{Taniai Benchmark}
	We evaluated PARN compared to other state-of-the-art methods on the Taniai benchmark \cite{Taniai16},
	which consists of 400 image pairs divided into three groups:
	FG3DCar, JODS, and PASCAL.
	Flow accuracy was
	measured by computing the proportion of foreground pixels with an
	absolute flow endpoint error that is smaller than a certain threshold
	$T$, after resizing images so that its larger dimension is 100
	pixels. 
	\figref{img:7} shows the flow accuracy with varying error threshold $T$.
	Our method outperforms especially when the error thershold is small. This clearly demonstrates the advantage of our hierarchical model in terms of both localization precision and appearance invariace. 
	
	\tabref{tab:1} summarizes the matching accuracy for various
	dense semantic correspondence techniques at the fixed threshold
	($T=5$ pixels). The quantitative results of `PARN-Lv1' and `CNNGM-Aff' in \tabref{tab:1} verify the benefits of our weakly supervised training scheme. Whereas `CNNGM-Aff.' is also trained in weakly supervised manner, it relys only on the synthetically deformed image pairs while our method employs semantically sensitive supervisions. Note that we implemented our regression module at level 1 in the same architecture of `CNNGM-Aff.'.
	From the qualitative results of \figref{img:8}, while DCTM is trapped in local minima unless an appropriate initial solution is given, 
	our method progressively predicts locally-varying
	affine transformation fields and able to handle relatively large semantic
	variations including flip variations without handcrafted parameter tuning. 
	The superiority of PARN can be seen by comparing to the correspondence techniques with the same `VGG-16' descriptor in \tabref{tab:1} and \figref{img:7} and even outperforms the supervised learning based method of \cite{han17}.
	We also evaluated with ResNet-101 \cite{He16} as a backbone network to demonstrate the performance boosting of our method with more powerful features, where our method achieves the best performance on average.\
	\vspace{-10pt}
	
	\begin{figure}[!t]
		\centering
		\renewcommand{\thesubfigure}{}
		\subfigure[]
		{\includegraphics[width=0.164\linewidth,clip]{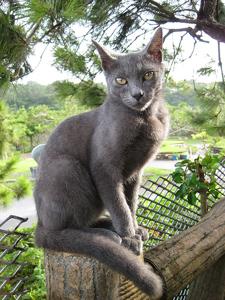}}\hfill
		\subfigure[]
		{\includegraphics[width=0.164\linewidth]{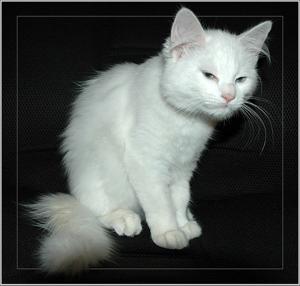}}\hfill
		\subfigure[]
		{\includegraphics[width=0.164\linewidth]{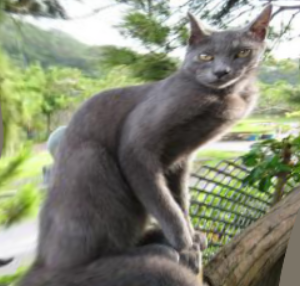}}\hfill
		\subfigure[]
		{\includegraphics[width=0.164\linewidth]{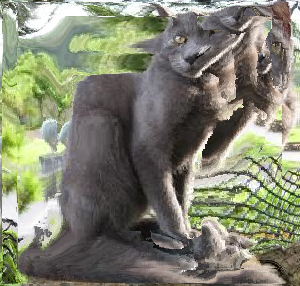}}\hfill
		\subfigure[]
		{\includegraphics[width=0.164\linewidth]{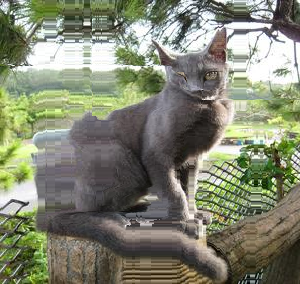}}\hfill
		\subfigure[]
		{\includegraphics[width=0.164\linewidth]{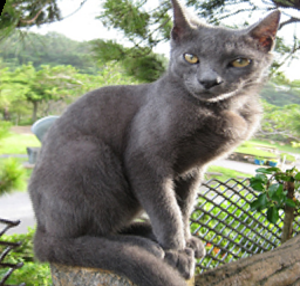}}\hfill
		\vspace{-21pt}
		\subfigure[(a)]
		{\includegraphics[width=0.164\linewidth,clip]{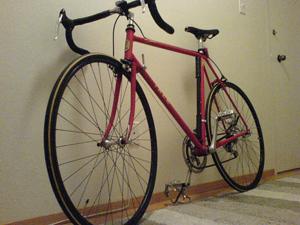}}\hfill
		\subfigure[(b)]
		{\includegraphics[width=0.164\linewidth]{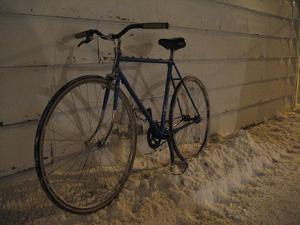}}\hfill
		\subfigure[(c)]
		{\includegraphics[width=0.164\linewidth]{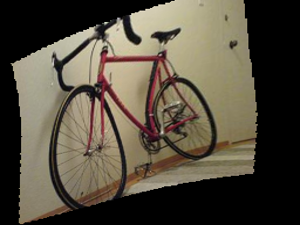}}\hfill
		\subfigure[(d)]
		{\includegraphics[width=0.164\linewidth]{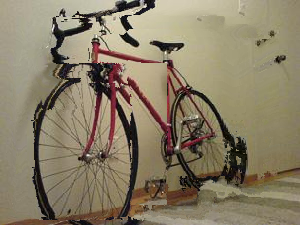}}\hfill
		\subfigure[(e)]
		{\includegraphics[width=0.164\linewidth]{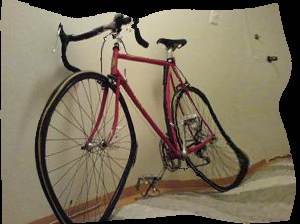}}\hfill
		\subfigure[(f)]
		{\includegraphics[width=0.164\linewidth]{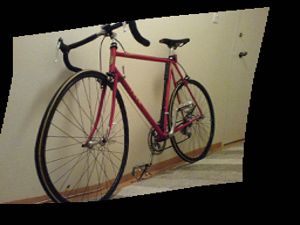}}\hfill
		\vspace{-10pt}
		\caption{Qualitative results on the PF-PASCAL benchmark \cite{Ham17}: (a) source image, (b) target image,
			(c) CNNGM-Aff.+TPS \cite{Rocco17}, (d) SCNet-AG+ \cite{han17}, (e) DCTM \cite{dctm}, (f) PARN.
			The source images were warped to the target images using correspondences.}\label{img:9}
	\end{figure}
	\begin{table}[t]
		\centering
		\begin{tabular}{ >{\raggedright}m{0.17\linewidth}
				>{\centering}m{0.13\linewidth} >{\centering}m{0.12\linewidth}
				>{\centering}m{0.13\linewidth}>{\centering}m{0.11\linewidth}  >{\centering}m{0.1\linewidth}
				>{\centering}m{0.11\linewidth}}
			\hlinewd{1.0pt}
			Dataset &\multicolumn{3}{c|}{PF-PASCAL} &\multicolumn{3}{ c }{Caltech-101}\tabularnewline
			\hline
			\multirow{2}{*}{Methods} &\multicolumn{3}{ c| }{PCK} &\multirow{2}{*}{LT-ACC} &\multirow{2}{*}{IoU} &\multirow{2}{*}{LOC-ERR}\tabularnewline
			\cline{2-4}
			&$\alpha=0.05$ &$\alpha=0.1$ &\multicolumn{1}{ c| }{$\alpha=0.15$}\tabularnewline
			\hline
			
			\hline
			SF\cite{Liu11} &0.192 &0.334 &\multicolumn{1}{ c| }{0.492} &0.75 &0.48 &0.32\tabularnewline
			DSP \cite{Kim13} &0.198 &0.372 &\multicolumn{1}{ c| }{0.414} &0.77 &0.47 &0.35\tabularnewline
			PF \cite{Ham16} &0.235 &0.453 &\multicolumn{1}{ c| }{0.621} &0.78 &0.50 &0.25\tabularnewline
			OHG \cite{yang2017object} & -  & -  &\multicolumn{1}{ c| }{ - } &0.81 &0.55 &0.19\tabularnewline
			\hline
			SCNet \cite{han17} &0.260 &0.482 &\multicolumn{1}{ c| }{0.658} &0.79 &0.51 &0.25\tabularnewline
			CNNGM \cite{Rocco17} &0.254 &0.461 &\multicolumn{1}{ c| }{0.641} &0.80 &0.56 &0.25\tabularnewline
			DCTM \cite{dctm} &0.257 &0.477 &\multicolumn{1}{ c| }{0.648} &0.84 &0.53 &\textbf{0.18}\tabularnewline
			\hline
			PARN  &\textbf{0.268} &\textbf{0.491} &\multicolumn{1}{ c| }{\textbf{0.662}} &\textbf{0.87} &\textbf{0.65} &0.21\tabularnewline
			\hlinewd{1.0pt}
		\end{tabular}\vspace{+6pt}
		\caption{Matching accuracy compared to state-of-the-art correspondence techniques on the PF-PASCAL benchmark \cite{Ham16} and Caltech-101 dataset \cite{li06}.}\label{tab:2}\vspace{-20pt}
	\end{table}
	\subsubsection{PF-PASCAL Benchmark}
	We also evaluated PARN on the testing set of PF-PASCAL benchmark \cite{Ham16}.
	For the evaluation metric, we used the probability of correct keypoint (PCK) between flow-warped keypoints and the ground-truth. The warped keypoints are deemed to be correctly predicted if they lie within
	$\alpha \cdot \mathrm{max}(h,w)$ pixels of the ground-truth keypoints for
	$\alpha \in [0,1]$, where $h$ and $w$ are the height and width of
	the object bounding box, respectively.
	\figref{img:9} shows qualitative results for dense flow estimation.
	
	Without ground-truth annotations, our PARN has shown the outperforming performance compared to other methods in \tabref{tab:2} where \cite{han17} is trained in fully supervised manner.
	The relatively modest gain may come from the limited evaluation only on the sparsely annotated keypoints of PF-PASCAL benchmark.
	However, the qualitative results of our method in \figref{img:9} indicates that the performance can be significantly boosted when dense annotations are given for evaluation.
	Although \cite{han17} estimates the sparse correspondences in a geometrically plausible model, they compute the final dense semantic flow by linearly interpolating them which may not consider the semantic structures of target image. 
	By contrast, our method leverages a pyramidal model where the smoothness constraint is naturally imposed among semantic scales within deep networks. 
	\vspace{-10pt}

	\subsubsection{Caltech-101 dataset}
	Our evaluations also include the Caltech-101 dataset \cite{li06}. Following the experimental protocal in \cite{li06}, we randomly selected 15 pairs of images for each object class, and evaluated matching accuracy with three metrics: label transfer accuracy (LT-ACC), the IoU metric, and the localization error (LOC-ERR) of corresponding pixel positions.
	Note that compared to other benchmarks described above, the Caltech-101 dataset provides image pairs from more diverse classes, enabling us to evaluate our method under more general correspondence settings.
	For the results, our PARN clearly outperforms the semantic correspondence techniques in terms of LT-ACC and IoU metrics.
	\tabref{tab:2} summarizes the matching accuracy compared to state-of-the-art methods.
	
	\section{Conclusion}\label{sec:5}
	We presented a novel CNN architecture, called PARN, which estimates locally-varying affine transformation fields across semantically similar images. Our method defined on pyramidal model first estimates a global affine transformation over an entire image and then progressively increases the transformation flexibility. In contrast to previous CNN based methods for geometric field estimations, our method yields locally-varying affine transformation fields that lie in the continuous solution space. Moreover, our network was trained in a weakly-supervised manner, using correspondence consistency within object bounding boxes in the training image pairs. 
	We believe PARN can potentially benefit instance-level object detection and segmentation, thanks to its robustness to severe geometric variations.
	\vspace{-10pt}
	
	\subsubsection{Acknowledment.}
	This research was supported by Next-Generation Information Computing Development Program through the National Research Foundation of Korea(NRF) funded by the Ministry of Science, ICT (NRF-2017M3C4A7069370).
	\vspace{-10pt}
	
	\begin{figure}[t!]
		\centering
		\renewcommand{\thesubfigure}{}
		\subfigure[(a)]
		{\includegraphics[width=0.164\linewidth]{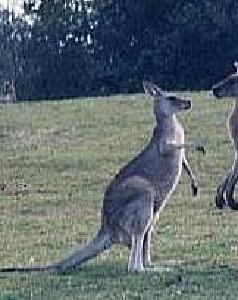}}\hfill
		\subfigure[(b)]
		{\includegraphics[width=0.164\linewidth]{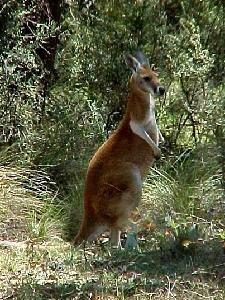}}\hfill
		\subfigure[(c)]
		{\includegraphics[width=0.164\linewidth]{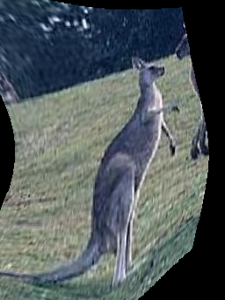}}\hfill
		\subfigure[(d)]
		{\includegraphics[width=0.164\linewidth]{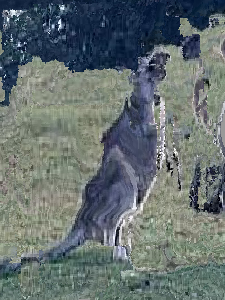}}\hfill
		\subfigure[(e)]
		{\includegraphics[width=0.164\linewidth]{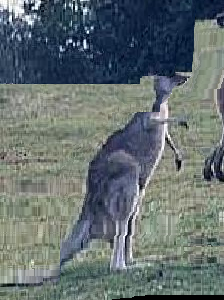}}\hfill
		\subfigure[(f)]
		{\includegraphics[width=0.164\linewidth]{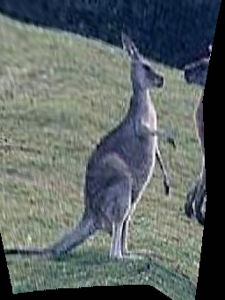}}\hfill
		\vspace{-10pt}
		\caption{Qualitative result on the Caltech-101 benchmark \cite{li06}: (a) source image, (b) target image,
			(c) CNNGM-Aff.+TPS\cite{Rocco17}, (d) SCNet-AG+ \cite{han17}, (e) DCTM \cite{dctm}, (f) PARN.
			The source image was warped to the target images using correspondences.}\label{img:10}\vspace{-10pt}
	\end{figure}
\bibliographystyle{splncs}
\bibliography{egbib}
\end{document}